\documentclass[final]{cvpr}

\usepackage{times}
\usepackage{epsfig}
\usepackage{graphicx}
\usepackage{amsmath}
\usepackage{amssymb}
\usepackage{tabu}
\usepackage{booktabs}
\usepackage{color}
\usepackage{breqn}
\usepackage{float}
\usepackage[numbers,sort&compress]{natbib}

\usepackage[pagebackref=true,breaklinks=true,colorlinks,bookmarks=false]{hyperref}

\def\cvprPaperID{5193} 
\def\confYear{CVPR 2021}

\begin{document}

\title{FineNet: Frame Interpolation and Enhancement for Face Video Deblurring}

\author{Phong Tran, Anh Tran, Thao Nguyen, Minh Hoai\\
VinAI Research\\
{\tt\small \{v.phongtt15, v.anhtt152, v.thaontp79\}@vinai.io, minhhoai@cs.stonybrook.edu}}

\maketitle

\def\mA{\mathcal{A}}
\def\mB{\mathcal{B}}
\def\mC{\mathcal{C}}
\def\mD{\mathcal{D}}
\def\mE{\mathcal{E}}
\def\mF{\mathcal{F}}
\def\mG{\mathcal{G}}
\def\mH{\mathcal{H}}
\def\mI{\mathcal{I}}
\def\mJ{\mathcal{J}}
\def\mK{\mathcal{K}}
\def\mL{\mathcal{L}}
\def\mM{\mathcal{M}}
\def\mN{\mathcal{N}}
\def\mO{\mathcal{O}}
\def\mP{\mathcal{P}}
\def\mQ{\mathcal{Q}}
\def\mR{\mathcal{R}}
\def\mS{\mathcal{S}}
\def\mT{\mathcal{T}}
\def\mU{\mathcal{U}}
\def\mV{\mathcal{V}}
\def\mW{\mathcal{W}}
\def\mX{\mathcal{X}}
\def\mY{\mathcal{Y}}
\def\mZ{\mathcal{Z}} 

\def\bbN{\mathbb{N}} 
\def\bbR{\mathbb{R}} 
\def\bbP{\mathbb{P}} 
\def\bbQ{\mathbb{Q}} 
\def\bbE{\mathbb{E}}

\def\1n{\mathbf{1}_n}
\def\0{\mathbf{0}}
\def\1{\mathbf{1}}

\def\A{{\bf A}}
\def\B{{\bf B}}
\def\C{{\bf C}}
\def\D{{\bf D}}
\def\E{{\bf E}}
\def\F{{\bf F}}
\def\G{{\bf G}}
\def\H{{\bf H}}
\def\I{{\bf I}}
\def\J{{\bf J}}
\def\K{{\bf K}}
\def\L{{\bf L}}
\def\M{{\bf M}}
\def\N{{\bf N}}
\def\O{{\bf O}}
\def\P{{\bf P}}
\def\Q{{\bf Q}}
\def\R{{\bf R}}
\def\S{{\bf S}}
\def\T{{\bf T}}
\def\U{{\bf U}}
\def\V{{\bf V}}
\def\W{{\bf W}}
\def\X{{\bf X}}
\def\Y{{\bf Y}}
\def\Z{{\bf Z}}

\def\a{{\bf a}}
\def\b{{\bf b}}
\def\c{{\bf c}}
\def\d{{\bf d}}
\def\e{{\bf e}}
\def\f{{\bf f}}
\def\g{{\bf g}}
\def\h{{\bf h}}
\def\i{{\bf i}}
\def\j{{\bf j}}
\def\k{{\bf k}}
\def\l{{\bf l}}
\def\m{{\bf m}}
\def\n{{\bf n}}
\def\o{{\bf o}}
\def\p{{\bf p}}
\def\q{{\bf q}}
\def\r{{\bf r}}
\def\s{{\bf s}}
\def\t{{\bf t}}
\def\u{{\bf u}}
\def\v{{\bf v}}
\def\w{{\bf w}}
\def\x{{\bf x}}
\def\y{{\bf y}}
\def\z{{\bf z}}

\def\balpha{\mbox{\boldmath{$\alpha$}}}
\def\bbeta{\mbox{\boldmath{$\beta$}}}
\def\bdelta{\mbox{\boldmath{$\delta$}}}
\def\bgamma{\mbox{\boldmath{$\gamma$}}}
\def\blambda{\mbox{\boldmath{$\lambda$}}}
\def\bsigma{\mbox{\boldmath{$\sigma$}}}
\def\btheta{\mbox{\boldmath{$\theta$}}}
\def\bomega{\mbox{\boldmath{$\omega$}}}
\def\bxi{\mbox{\boldmath{$\xi$}}}
\def\bnu{\mbox{\boldmath{$\nu$}}}                                  
\def\bphi{\mbox{\boldmath{$\phi$}}}
\def\bmu{\mbox{\boldmath{$\mu$}}}

\def\bDelta{\mbox{\boldmath{$\Delta$}}}
\def\bOmega{\mbox{\boldmath{$\Omega$}}}
\def\bPhi{\mbox{\boldmath{$\Phi$}}}
\def\bLambda{\mbox{\boldmath{$\Lambda$}}}
\def\bSigma{\mbox{\boldmath{$\Sigma$}}}
\def\bGamma{\mbox{\boldmath{$\Gamma$}}}
                                  
\newcommand{\myprob}[1]{\mathop{\mathbb{P}}_{#1}}

\newcommand{\myexp}[1]{\mathop{\mathbb{E}}_{#1}}

\newcommand{\mydelta}[1]{1_{#1}}

\newcommand{\myminimum}[1]{\mathop{\textrm{minimum}}_{#1}}
\newcommand{\mymaximum}[1]{\mathop{\textrm{maximum}}_{#1}}    
\newcommand{\mymin}[1]{\mathop{\textrm{minimize}}_{#1}}
\newcommand{\mymax}[1]{\mathop{\textrm{maximize}}_{#1}}
\newcommand{\mymins}[1]{\mathop{\textrm{min.}}_{#1}}
\newcommand{\mymaxs}[1]{\mathop{\textrm{max.}}_{#1}}  
\newcommand{\myargmin}[1]{\mathop{\textrm{argmin}}_{#1}} 
\newcommand{\myargmax}[1]{\mathop{\textrm{argmax}}_{#1}} 
\newcommand{\myst}{\textrm{s.t. }}

\newcommand{\denselist}{\itemsep -1pt}
\newcommand{\sparselist}{\itemsep 1pt}

\definecolor{pink}{rgb}{0.9,0.5,0.5}
\definecolor{purple}{rgb}{0.5, 0.4, 0.8}   
\definecolor{gray}{rgb}{0.3, 0.3, 0.3}
\definecolor{mygreen}{rgb}{0.2, 0.6, 0.2}

\newcommand{\cyan}[1]{\textcolor{cyan}{#1}}
\newcommand{\red}[1]{\textcolor{red}{#1}}  
\newcommand{\blue}[1]{\textcolor{blue}{#1}}
\newcommand{\magenta}[1]{\textcolor{magenta}{#1}}
\newcommand{\pink}[1]{\textcolor{pink}{#1}}
\newcommand{\green}[1]{\textcolor{green}{#1}} 
\newcommand{\gray}[1]{\textcolor{gray}{#1}}    
\newcommand{\mygreen}[1]{\textcolor{mygreen}{#1}}    
\newcommand{\purple}[1]{\textcolor{purple}{#1}}       

\definecolor{greena}{rgb}{0.4, 0.5, 0.1}
\newcommand{\greena}[1]{\textcolor{greena}{#1}}

\definecolor{bluea}{rgb}{0, 0.4, 0.6}
\newcommand{\bluea}[1]{\textcolor{bluea}{#1}}
\definecolor{reda}{rgb}{0.6, 0.2, 0.1}
\newcommand{\reda}[1]{\textcolor{reda}{#1}}

\def\changemargin#1#2{\list{}{\rightmargin#2\leftmargin#1}\item[]}
\let\endchangemargin=\endlist
                                               
\newcommand{\cm}[1]{}

\newcommand{\mhoai}[1]{{\color{purple}\textbf{[MH: #1]}}}

\newcommand{\mtodo}[1]{{\color{red}$\blacksquare$\textbf{[TODO: #1]}}}
\newcommand{\myheading}[1]{\vspace{1ex}\noindent \textbf{#1}}
\newcommand{\htimesw}[2]{\mbox{$#1$$\times$$#2$}}


\newif\ifshowsolution
\showsolutiontrue

\ifshowsolution  
\newcommand{\Comment}[1]{\paragraph{\bf $\bigstar $ COMMENT:} {\sf #1} \bigskip}
\newcommand{\Solution}[2]{\paragraph{\bf $\bigstar $ SOLUTION:} {\sf #2} }
\newcommand{\Mistake}[2]{\paragraph{\bf $\blacksquare$ COMMON MISTAKE #1:} {\sf #2} \bigskip}
\else
\newcommand{\Solution}[2]{\vspace{#1}}
\fi

\newcommand{\truefalse}{
\begin{enumerate}
	\item True
	\item False
\end{enumerate}
}

\newcommand{\yesno}{
\begin{enumerate}
	\item Yes
	\item No
\end{enumerate}
}

\newcommand{\mh}[1]{\textcolor{blue}{[Hoai: {#1}]}}
\newcommand{\at}[1]{\textcolor{red}{[Anh: {#1}]}}
\newcommand{\fong}[1]{\textcolor{blue}{[Phong: {#1}]}}
\newcommand{\Sref}[1]{Sec.~\ref{#1}}
\newcommand{\Eref}[1]{Eq.~(\ref{#1})}
\newcommand{\Fref}[1]{Fig.~\ref{#1}}
\newcommand{\Tref}[1]{Table~\ref{#1}}
\newcommand{\cellimg}[1]{
    \includegraphics[width=1.7cm, height=1.7cm]{#1}
}
\newcommand{\cellimgsmall}[1]{
    \includegraphics[width=1.5cm, height=1.7cm]{#1}
}

\begin{abstract}
   The objective of this work is to deblur face videos. We propose a method that tackles this problem from two directions: (1) enhancing the blurry frames, and (2) treating the blurry frames as missing values and estimate them by interpolation. These approaches are complementary to each other, and their combination outperforms individual ones. We also introduce a novel module that leverages the structure of faces for finding positional offsets between video frames. This module can be integrated into the processing pipelines of both approaches, improving the quality of the final outcome. Experiments on three real and synthetically generated blurry video datasets show that our method outperforms the previous state-of-the-art methods by a large margin in terms of both quantitative and qualitative results.
\end{abstract}

\section{Introduction}

Many face videos contain blurry frames, due to the amateur use of hand-held cameras or the fast movement of highly animated faces. Blurry faces are very unpleasant to watch, and they are the causes of failure for downstream tasks such as gaze estimation and expression recognition. Thus face deblurring is important for many applications. 

Deblurring is an important research problem, and it has been extensively studied in the literature of signal processing and computer vision. Recent state-of-the-art methods \cite{Ren-ICCV-2019,caballero2017real,tao2017detail,wang2019edvr,tian2018tdan,jo2018deep} are based on deep learning, where a convolutional neural network is trained to enhance the blurry frames by exploiting the temporal redundancy among neighboring frames. A key step in the processing pipeline of these methods is \textsl{alignment}, where the neighboring frames are warped to align with the target frame in consideration~\cite{caballero2017real,xue2019video,kim2018spatio}. Unfortunately, alignment can be difficult if the target frame is too blurry, and inaccurate alignment leads to poor deblurring results. 

\begin{figure}[t]
\centering
\includegraphics[width=0.9\linewidth]{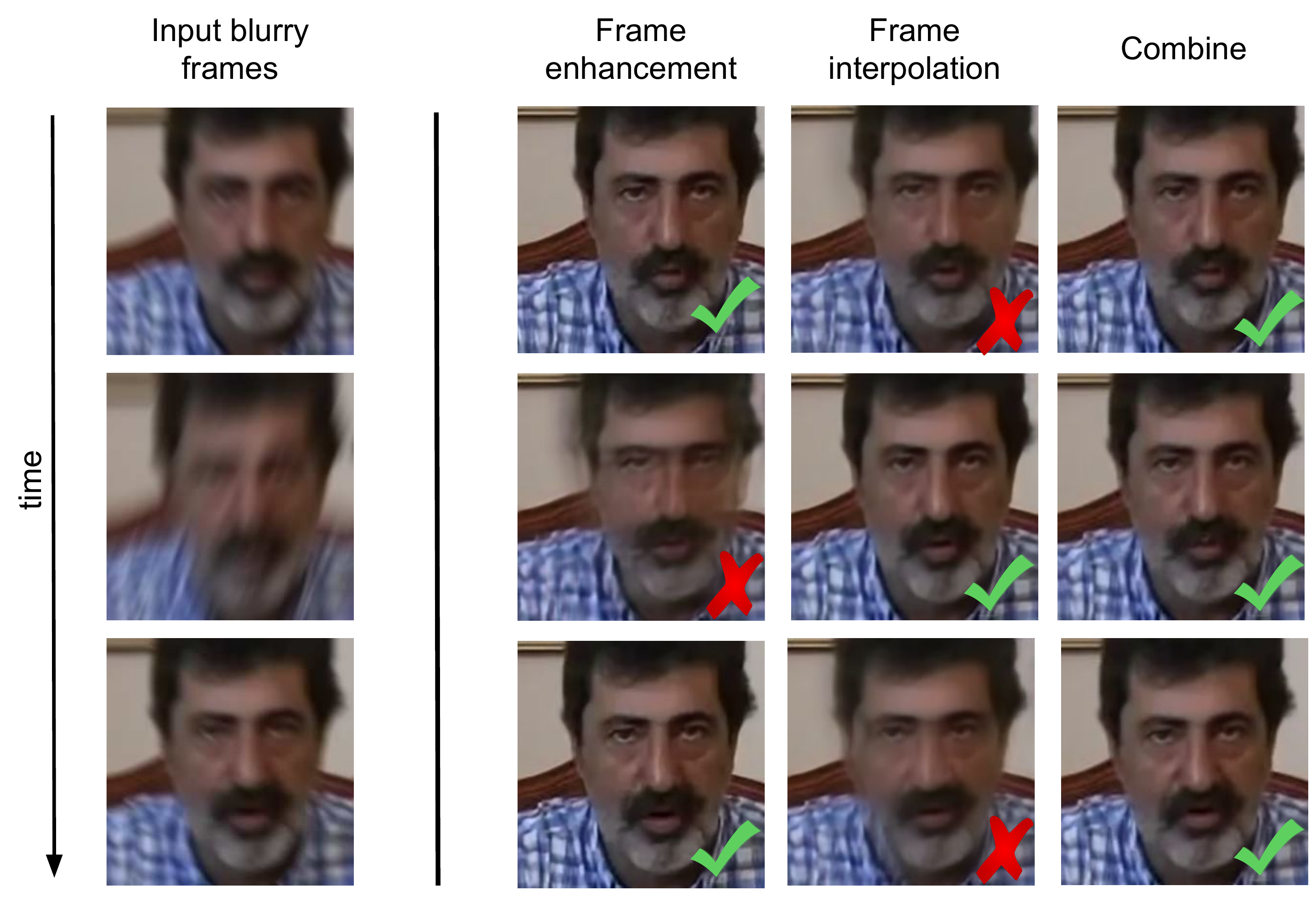}
%
\caption{We propose to combine Frame Enhancement and Frame Interpolation for face debluring. These approaches are complementary, with their own pros and cons. Green checkmarks and red crosses indicate high and low quality images. On the severely blurry Frame~2, Interpolation is better than Enhancement. On less blurry frames, Enhancement is better. The combination of Interpolation and Enhancement yields the highest quality.\label{fig:teaser}}
\end{figure}

\begin{figure}[t]
    \centering
    \includegraphics[width=0.8\linewidth]{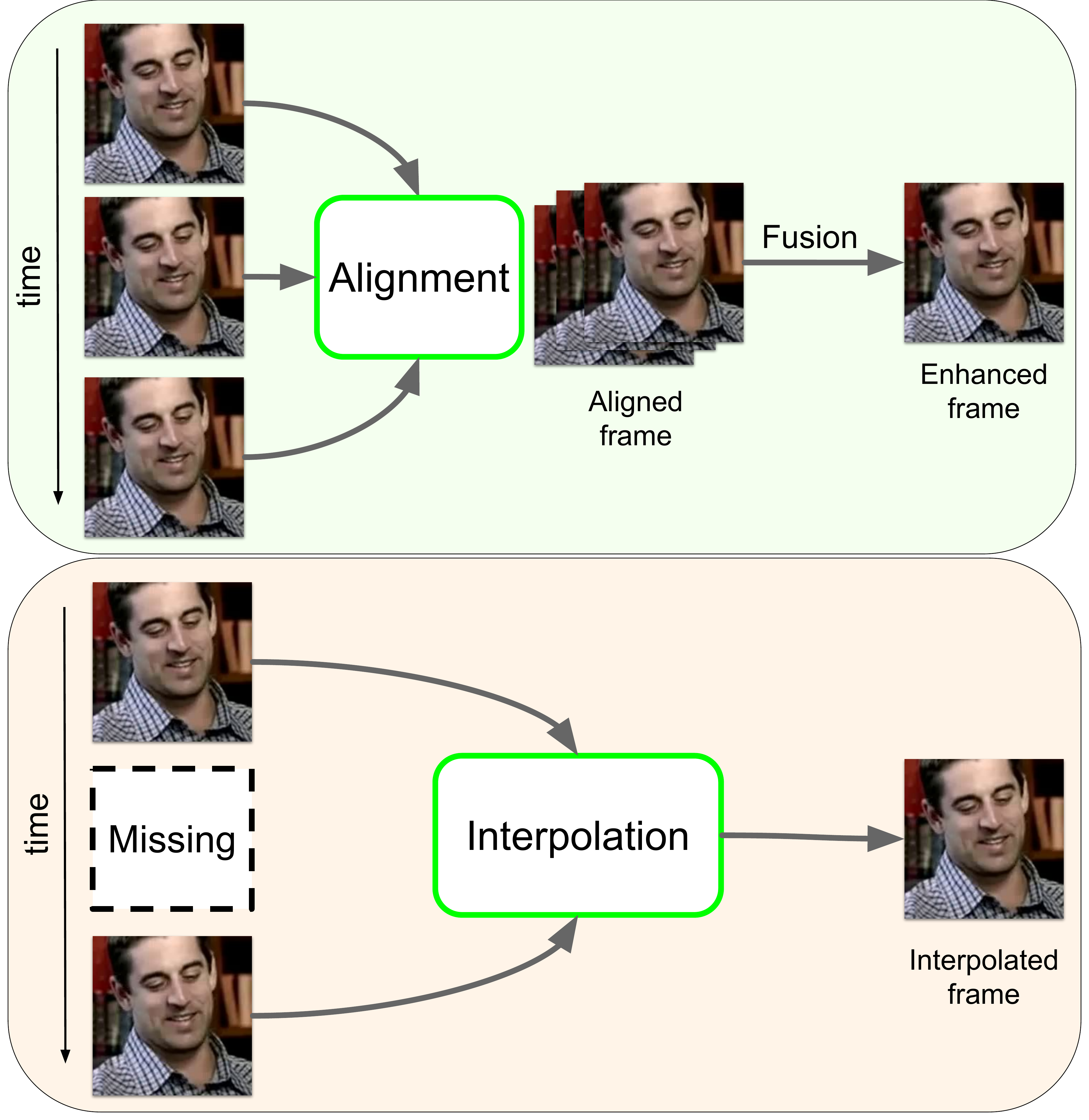}
    \caption{{\bf Different directions to tackle the deblurring problem}. Top: the enhancement approach, where the blurry target frame is enhanced with the assistance from the warped neighboring frames. Bottom: the interpolation approach, where the blurry frame is treated as a missing frame. \label{fig::semantic}}
\end{figure}

In this paper, we propose to tackle deblurring from an additional direction.  In addition to treating deblurring as an enhancement problem, we consider a blurry frame as a missing frame and estimate it by interpolating from neighboring frames. We refer to this novel approach as frame \textsl{interpolation}, which is radically different from the popular frame \textsl{enhancement} approach~\cite{caballero2017real,xue2019video,kim2018spatio}, as illustrated in \Fref{fig::semantic}. One advantage of the interpolation approach is that it bypasses the problematic blurry frame, while the enhancement approach has to use the blurry frame as the alignment target. On the other hand, the enhancement approach does have advantages over the interpolation approach,  especially when the target frame is already of high quality, so it is easier to enhance than to `create' from scratch. Given the complementary benefits of these two approaches, we propose to combine them to improve the quality of the deblurred output video. We will refer to our method as FineNet, which stands for \underline{F}rame \underline{I}nterpolation a\underline{N}d \underline{E}nhancement \underline{Net}work. The benefits of FineNet over individual approaches are illustrated in \Fref{fig:teaser}.



In addition to tackling deblurring from two directions, another technical novelty of our paper is the development of the Face-Aware Offset Calculation (FOC)  modules for frame alignment and interpolation. The idea is to use facial landmarks as a prior for finding correspondences and calculating positional offsets between image frames, and the facial landmarks can be obtained by running a facial landmark detector. However, the facial landmark detector might fail catastrophically, and it is challenging to design a robust approach to use facial landmarks. In this work, instead of defining explicit constraints based on the predicted landmarks, we use the landmark heatmaps to estimate the positional offsets for the deformable convolution kernels \cite{dai2017deformable,zhu2019deformable}. This approach is robust to the inaccuracy of the detected facial landmarks, and it improves the quality of the final deblurring outputs. 

We compare the  proposed method with other state-of-the-art deblurring methods on several popular facial video benchmark datasets and find that our model quantitatively outperforms the others by a large margin. We also evaluate on Internet videos with real blur and observe that our model produces clearer and sharper results.


\section{Related work}
\myheading{Image and video deblurring.}
Traditional deblurring methods, deconvoloution, aim to recover from a given blurry image $y$ the original sharp image $x$ and the blur kernel $k$ (usually assumed to be a convolution kernel). This is an ill-posed task, and it requires some prior knowledge of the sharp image and the blur kernel. \citet{chan1998total} suggested using a total variation penalty to enforce the sparsity of the sharp image gradient. \citet{pan2016blind} discovered that the dark channel of the sharp image was sparser than in the blurry one, thus proposed a dark channel prior favoring sharp image in a MAP-based optimization method. \citet{liu2014blind} and \citet{zuo2016learning} proposed some common priors for the blur kernel, but these methods are computationally expensive. Furthermore, the proposed priors may not be suitable to in-the-wild blurry images. Most traditional algorithms struggle with non-linear and non-uniform blur kernels.

Together with the advancement of deep learning, there has been tremendous improvement in the performance of deblurring methods in the last few years. Video deblurring was initially considered as a special case of image deblurring, but many recent video deblurring methods also used temporal information to improve performance. One popular approach is to align neighboring frames by using motion field and subsequently fuse the aligned frames to recover the sharp image. Optical flow was used for frame warping in \cite{Ren-ICCV-2019,caballero2017real,tao2017detail}, while neighboring frames were used to estimate kenrels of a convolutional network \cite{wang2019edvr,tian2018tdan,jo2018deep}. In this paper, we will furthermore exploit the temporal information by a frame interpolation module.

\myheading{Facial image enhancement.}
Face deblurring can improve the performance of various face analysis systems, including gaze estimation, expression analysis, and 3D face modeling. In many selfie and streaming applications such as Skype and Zoom, faces are often the main focus of a video, so a face-specific deblurring algorithm is preferred over a generic method. To this end, the proposed face-specific deblurring method is well aligned with a number of image enhancement techniques that have been developed specifically for faces, exploiting the semantic structures of faces. The facial structures can be used as priors for face deblurring. One common prior is semantic segmentation.  \citet{shen2018deep} concatenated face semantic labels with face images as the input of the model. \citet{yasarla2019deblurring} parsed a face into four parts, deblured each part separately before combining the results. Beyond 2D priors, 3D face rendering has also been used to guide the deblurring process \cite{Ren-ICCV-2019}. One common problem of the aforementioned methods is that they rely heavily on the facial estimation modules, but face parsing and 3D face rendering modules do not work well with blurry faces. That explains why the deblurring algorithms that explicitly and directly use face priors cannot handle severe blur. To avoid the problem of existing methods, we use facial landmarks to learn the dynamic kernels of deformable convolution layers, and this approach is more robust to the failure of the facial landmark detection module. 


\myheading{Using facial landmarks for deblurring.}
Thanks to deep learning, facial landmark detection has had a great leap in performance lately. Landmark detection methods \cite{bulat2017far,bulat2017binarized,yang2017stacked,dong2018style,valle2019face} can now produce reasonable facial key-points even on blurry images.
Most current state-of-the-art methods  \cite{bulat2017far,yang2017stacked,tang2018quantized,bulat2017binarized} use CNNs to regress a heatmap for each landmark point and subsequently use those heatmaps to generate the locations of landmarks. In this paper, we use those intermediate heatmaps instead of the predicted points. 

\citet{sun2019fab} proposed a cyclic process to connect deblurring and landmarks detection. Their model included two branches corresponding to the two mentioned tasks. The output of the first branch was used as the input of the second branch. The output of the second branch, in turn, was used as a prior for the first branch. However, this process was overly time-consuming. Following \citet{wang2019edvr}, we split the process into two stages. At the first stage, we use a smaller FineNet to remove noise and reduce blur density. At the second stage, we apply our full model.

\begin{figure*}[t]
\begin{center}
\includegraphics[width=0.7\linewidth]{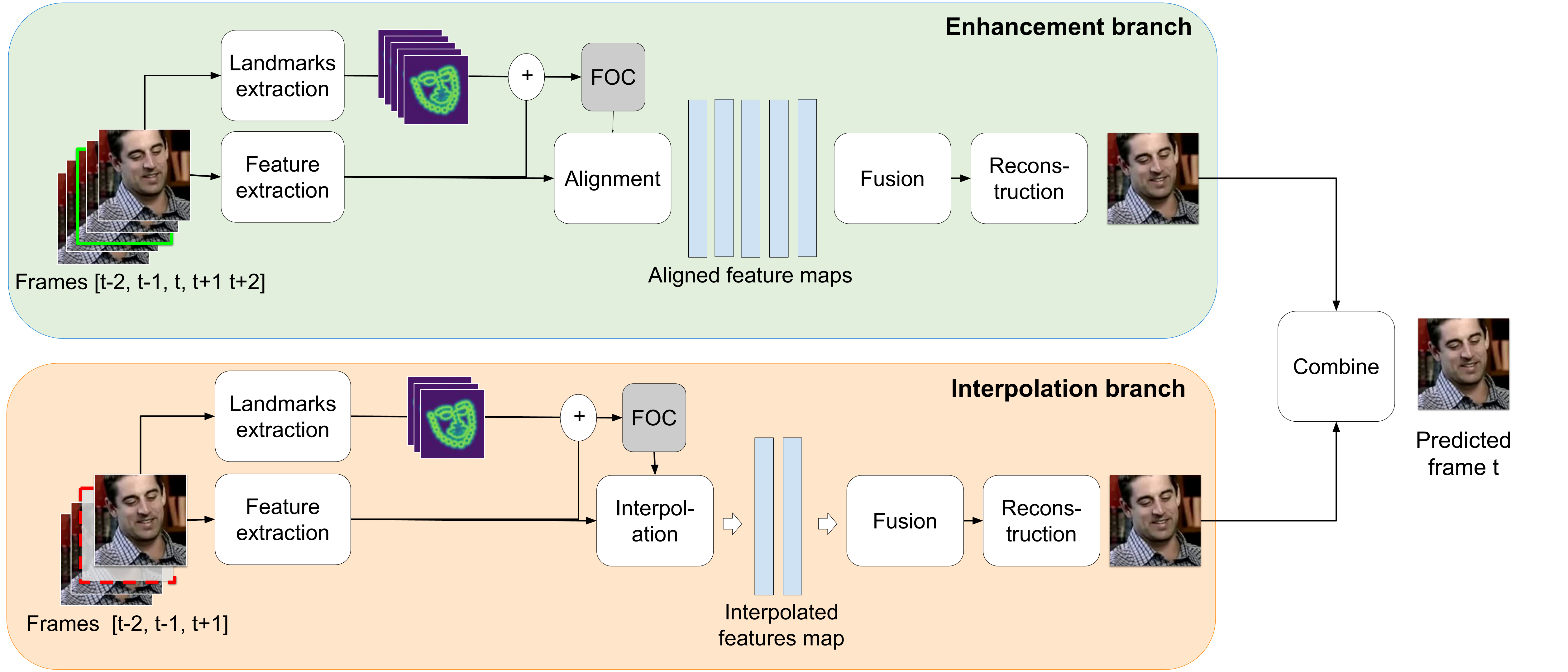}
\caption{General architecture of our model \label{fig:twobranch}}
\end{center}
\end{figure*}

\section{FineNet}
FineNet has two branches: enhancement and interpolation, as illustrated in \Fref{fig:twobranch}. The enhancement branch aims to enhance each blurry frame directly, with the assistance of warped frames by aligning the nearby frames to the target frame. The interpolation branch treats the blurry frame as a missing observation and estimates it from the nearby frames. The output images from the enhancement and interpolation branches are fused to produce the final output. 

To improve both the enhancement and interpolation branches,  we leverage the facial landmarks to calculate the location offsets for the deformable convolutional kernels~\cite{dai2017deformable, zhu2019deformable}. We introduce Face-aware Offset Calculation (FOC), a novel module that takes the target frame, the neighboring frames, and their facial landmark heatmaps to predict the alignment offsets. FOC modules will be used in both the enhancement and interpolation branches.
For a better understanding of the proposed method, we first review deformable convolution and the important roles of location offsets. We will then describe FOC, followed by the description of the enhancement and interpolation branches.

\subsection{Preliminaries: deformable convolution} Deformable Convolution Network (DCN) was proposed by \citet{dai2017deformable}. Instead of using convolution kernels with rigid and uniform receptive fields, DCN performs convolution operators on deformable locations with the location offsets being specified by an additional network branch. \citet{zhu2019deformable} extend DCN by introducing additional learnable modulation scalars $\Delta m$ to improve the performance of deformable convolution operation. Deformable convolution has been used in the alignment module of deblurring methods \cite{wang2019edvr,tian2018tdan}, yielding impressive results. Inspired by their success, we also propose to use deformable convolution in both the enhancement and interpolation branches. 

Given a feature map $F$, the feature at position $p$ in the next layer $F'(p)$ can be obtained by a convolution layer with a kernel of size $3{\times}3$ as follows:
\begin{equation}
    F'(p) = \sum_{k=1}^{9}w_k * F(p + p_k).
\end{equation}
where $p_k \in \{(-1, -1), (-1, 0), \cdots \}$ and $w_k$ is the weight of the location $p_k$ in the kernel. Different from normal convolution network, modulated deformable convolution network \cite{zhu2019deformable} has two additional learnable offsets $\Delta p_k$ and modulation scalars $\Delta m_k$. Now the feature at position $p$ in the next layer $F'(p)$ is:
\begin{equation}
    F'(p) = \sum_{k=1}^{9}w_k * F(p + p_k + \Delta p_k) * \Delta m_k.
\end{equation}

\begin{figure}[t]
\centering
\includegraphics[width=0.88\linewidth]{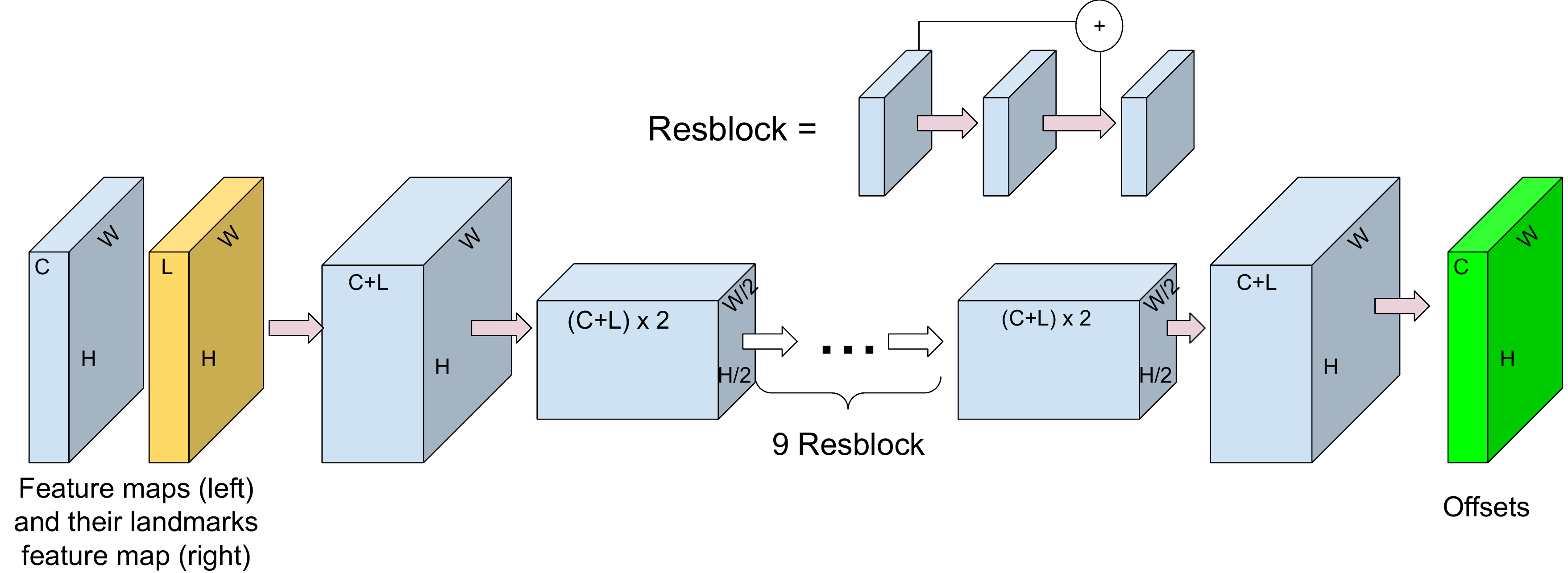}
\caption{{Architecture of a Face-aware Offset Calculation (FOC) module.} \label{fig:fog}}
\end{figure}

\subsection{Face-aware Offset Calculation (FOC) \label{sec:fog}}
Offset estimation is a critical step in the DCN-based feature alignment, but it was overlooked in previous deblurring works \cite{wang2019edvr,tian2018tdan}, which used only two convolution layers for offset estimation. We instead design a dedicated module called Face-aware Offset Calculation (FOC). A FOC module takes features and landmark heatmaps of several frames as input. The output of the module is an offset map for alignment or interpolation. More specifically, suppose we have a set of frames and the corresponding extracted heatmaps (using~\cite{bulat2017far}). We use a feature extraction module to convert them to feature maps, denoted as $\mathcal{F} = \{F_0, \cdots, F_k\}$ and $\mathcal{H} = \{H_0, \cdots, H_k\}$ respectively. Then, we concatenate $\mathcal{F}$ and $\mathcal{H}$ together and put them into a FOC module to get the offset. In the enhancement branch, a FOC module takes one neighboring frame and the target frame as input. A deformable convolution layer will use the estimated offset values to align the neighboring feature map to the target one. In the interpolation branch, a FOC module takes three input frames either $[t - 2, t - 1, t + 1]$ or $[t + 2, t + 1, t - 1]$. A deformable convolution layer will take the computed offset values to convert the middle frame of the input set to the target frame $t$. The architecture of a FOC module is illustrated in \Fref{fig:fog}.


The set of face landmark heatmaps is a crucial component in FOCs. Under severe blur, landmark heatmaps can be inaccurate and affect the performance of FOCs. However, we only use these keypoints to support the alignment step, while other papers use them to guide the reconstruction module directly. Our approach reduces the effect of incorrect landmark detection. Furthermore, we combine multiple alignments in the fusion step, making the system more robust to landmark localization errors. 


\begin{figure}[t]
    \centering
    \includegraphics[width=0.7\linewidth]{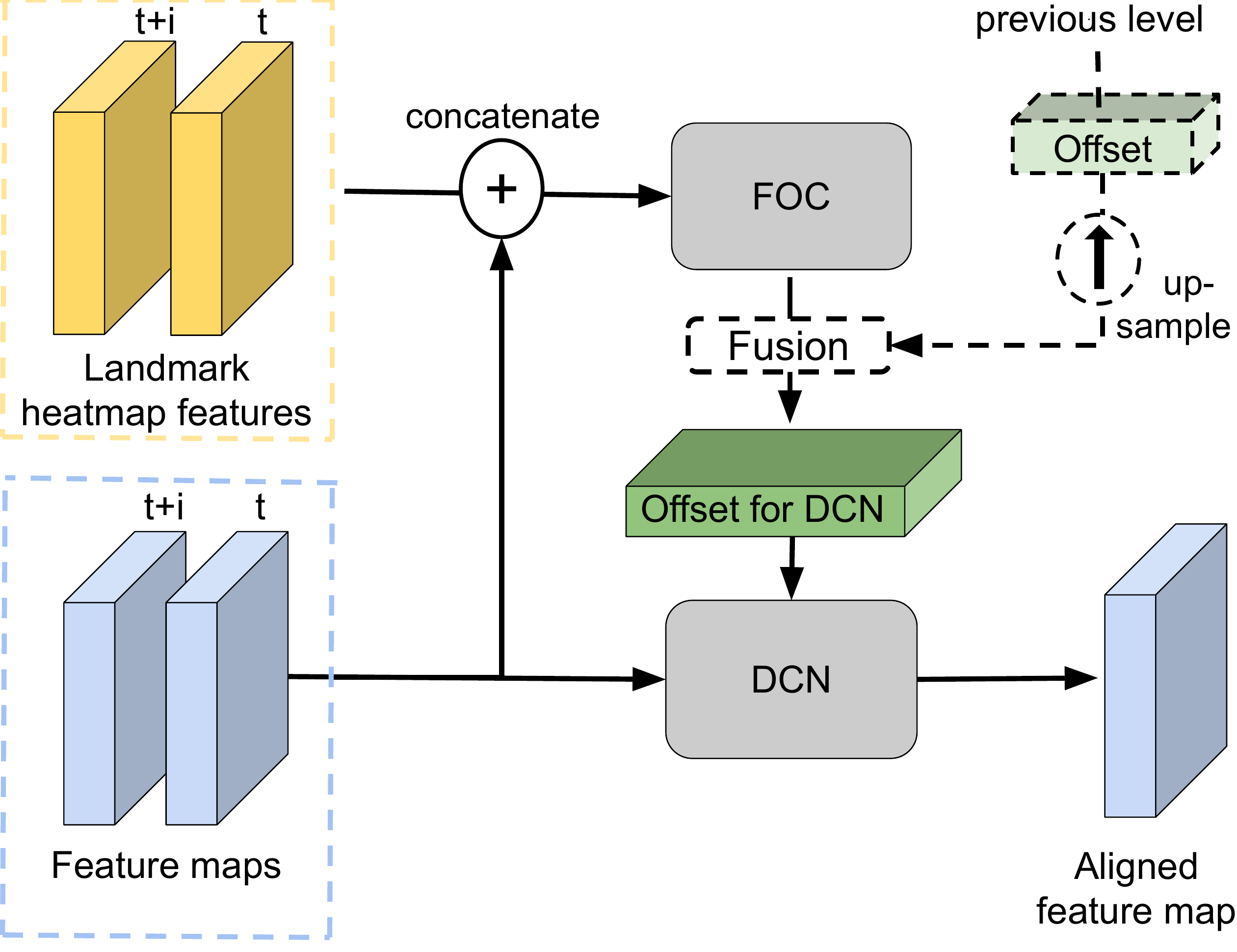} \hspace{2mm}
    \caption{\bf Single level alignment module \label{fig:singlePCD}}
\end{figure}

\begin{figure}[t]
    \centering
    \includegraphics[width=0.7\linewidth]{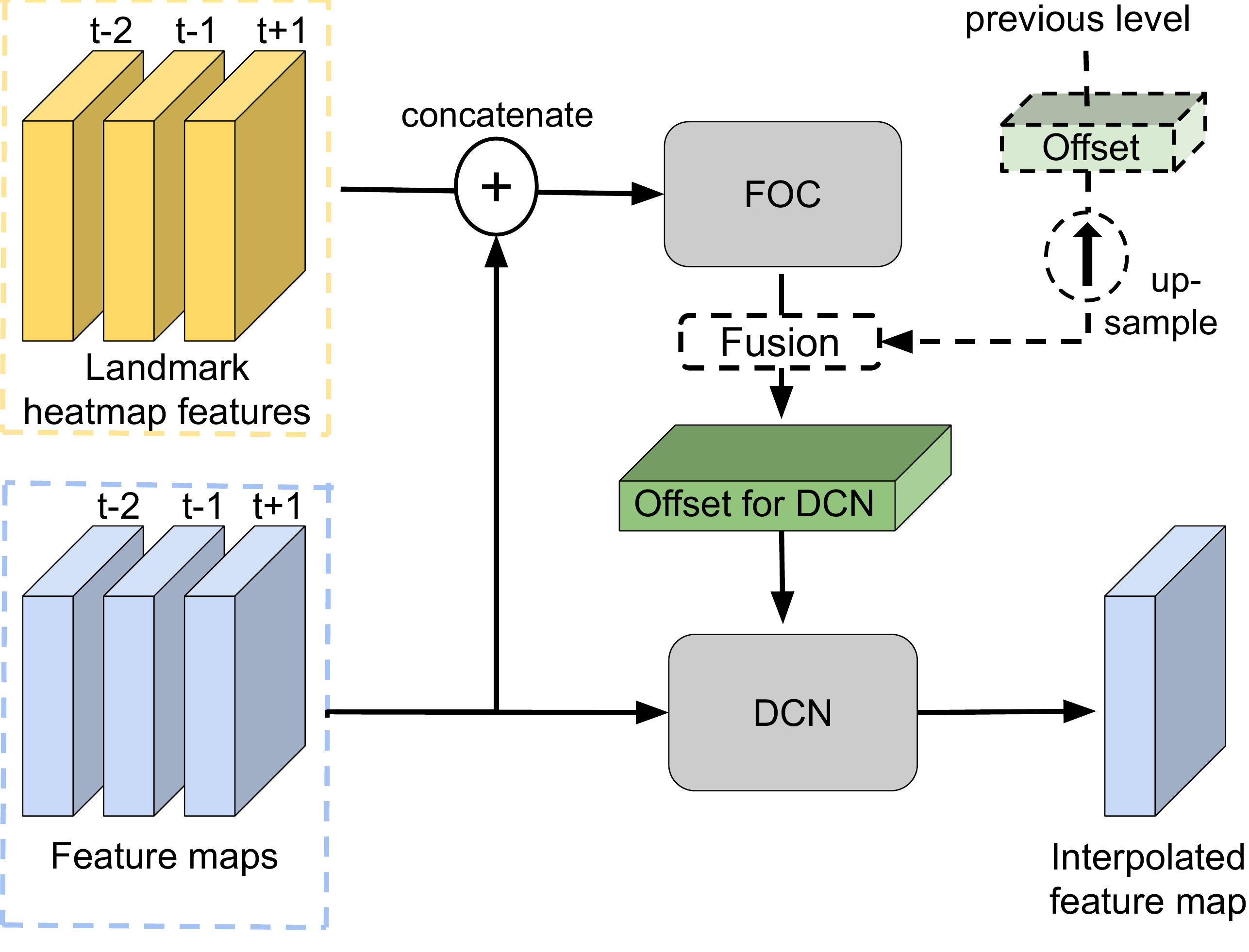} \hspace{2mm}
    \caption{\bf Single level interpolation module \label{fig:singleGMF}}
\end{figure}

Hereafter, for brevity while describing the network branches, we will use the terms ``frame'' and ``heatmap'' to imply the corresponding feature maps. 

\subsection{Enhancement branch}

To refine a blurry frame $t$, the enhancement branch uses information from five consecutive frames $t - 2$, $t-1$, $t$, $t+1$, $t + 2$. This branch has four processing steps: features extraction, alignment, fusion, and reconstruction. Each frame after downsampling goes through the features extraction module to generate a feature map. The feature maps of all frames are aligned to the feature map of frame $t$ using the alignment module. The alignment module is an extension of the PCD alignment module \cite{wang2019edvr} with our proposed FOC module. All the aligned feature maps, including the feature map of the target frame $t$, are combined by the fusion module to estimate the feature map of the enhanced target frame. Finally, the predicted target frame can be generated by its feature map using the reconstruction module.

We develop the alignment module based on the PCD align module \cite{wang2019edvr}, but improve it by using a FOC module to calculate the offsets for the DCN kernels. As will be seen in our experiments, this extension by itself already yields a model that outperforms the previous state-of-the-art deblurring methods. 

 
To align frame $t + i$ to frame $t$, we compute offsets and aligned features at multiple scales from coarse to fine as suggested by \citet{wang2019edvr}. For each scale level, we first use the feature maps and landmark heatmaps of the downsampled images to calculate the deformable convolution kernels as described in \Sref{sec:fog}. Except for the bottom level, the kernels will be refined by fusing with the upsampled offset estimation from the previous level. Finally, we convolve those deformable kernels with the feature map of frame $t+i$. \Fref{fig:singlePCD} depicts this alignment procedure.

Following~\cite{wang2019edvr}, we use a pyramid and cascading structure for alignment, as can be seen in \Fref{fig:PCD}. For each frame, we compute a pyramid of feature maps. The first level of the pyramid is the feature map extracted by a feature extraction module. From the second level, the feature map of the pyramid will be calculated directly from the previous level by a stride convolutional operation. The number of levels used in this paper is three. We index the top level as Level 1, and the bottom one is Level 3. To align frame $t' = t + i$ to frame $t$, each level in the pyramid structure of frame $t'$ is aligned with that respective level in the pyramid of frame $t$. At each non-leaf level $l-1$, we refine the aligned features by concatenating them with the corresponding upsampled features from level $l$, followed by a convolutional layer. For offset calculation, we do the same but replace the convolutional layer by a FOC module. Formally, let $F_{t}^l$ and $F_{t'}^l$ be the feature maps of frame $t$ and $t'$ at level $l$, $A_{t'}^l$ the desired aligned feature map and $\Delta P_{t'}^l$ the map for location offsets. Also, we denote $H_{t}$ and $H_{t'}$ as the landmark heatmaps of frame $t$ and $t'$. $A_{t'}^l$ and $\Delta P_{t'}^l$ are calculated as follows: 
\begin{align}
&    \Delta P_{t'}^3 = FOC(F_{t'}^3, F_{t}^3), \\
&    A_{t'}^3 = DConv(F_{t'}^3, \Delta P_{t'}^3)\\
&    \Delta P_{t'}^2 = FOC(Conv(F_{t'}^2 \oplus F_{t}^2), (\Delta P_{t'} ^{3})^{\uparrow 2}),\label{eqn:eq5}\\
&    A_{t'}^2 = Conv(DConv(F_{t'}^2, \Delta P_{t'}^2) \oplus (A_{t'}^{3})^{\uparrow 2}), \label{eqn:eq6}\\
&    \Delta P_{t'}^1 = FOC(Conv(F_{t'}^1 \oplus F_{t}^1), (\Delta P_{t'} ^{2})^{\uparrow 2}, H_{t'}, H_t),\label{eqn:eq7}\\
&    A_{t'}^1 = Conv(DConv(F_{t'}^1, \Delta P_{t'}^1) \oplus (A_{t'}^{2})^{\uparrow 2}), \label{eqn:eq8}
\end{align}
where $\oplus$ and $\uparrow$ are the concatenation and upsampling operators.
We only use facial landmark heatmaps in the FOC module of Level 0. When aligning feature maps of each non-top level, we use a simplified FOC module with a small encoder-decoder network that does not use the landmark prior to save computational cost.


\iftrue
\begin{figure*}[t]
\centering
\includegraphics[width=0.7\linewidth]{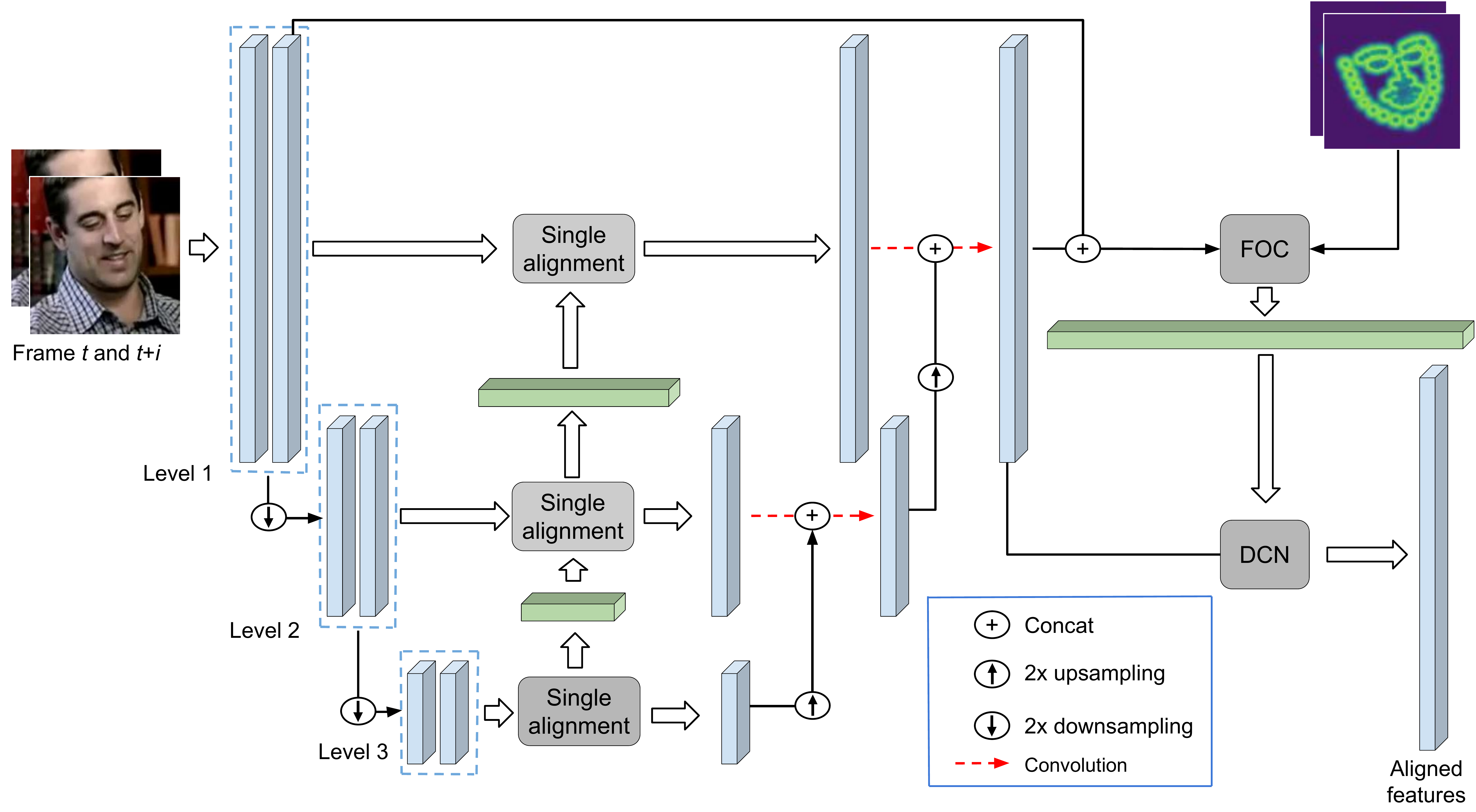}
\caption{Pyramid and cascading structure of the enhancement branch}
\label{fig:PCD}
\vspace{-4mm}
\end{figure*}
\fi

\subsection{Interpolation branch}

The interpolation branch also uses deformable convolution with a feature pyramid and cascading architecture. The `missing' frame $F_t$ is interpolated from  $F_{t - 2}, F_{t - 1}$, $F_{t + 1}, F_{t+2}$ and their corresponding landmark heatmaps $H_{t-2}$, $H_{t-1}$, $H_{t+1}$, $H_{t+2}$. We use both forward and backward interpolations, denoted by $F_t^{\rightarrow}$ and $F_t^{\leftarrow}$ respectively, as follows. First, we compute $F_t^{\rightarrow}$ by applying the interpolation module on three frames $[F_{t - 2}, F_{t - 1}, F_{t + 1}]$ and their landmark heatmaps. Second, we compute $F_t^{\leftarrow}$ by applying the interpolation module on three frames $[F_{t + 2}, F_{t + 1}, F_{t - 1}]$ and their landmark heatmaps (note the decreasing order of the frames). Finally, we combine $F_t^{\rightarrow}$ and $F_t^{\leftarrow}$ by a fusion module and predict the missing frame $F_t$ using the combined feature map. \Fref{fig:singleGMF} depicts the forward interpolation procedure. 



For both forward and backward interpolation, we use three neighboring frames instead of four. In our ablation studies described in the experiment section, there is no benefit of using four frames over three, so we opt to use three frames to reduce computational cost. 

Note that our frame interpolation module is designed specifically for the deblurring purposes. While this method may be applicable to a similar application of video slow-motion, it is beyond the scope of our paper.

Although two branches use a very similar network architecture, they perform two very different tasks. In the enhancement branch, each neighbor frame is aligned to the target frame separately, then fused together. This branch heavily relies on the target frame and aims to enhance it. In the interpolation branch, we ignore the target blurry frame and recover it from its neighbors.

One may question if the interpolation branch is effective when consecutive frames are blurry. We analyzed the real blurry videos and found interpolation useful most of the time. In many cases, the number of consecutive blurry frames is small such as abrupt face movement or low FPS videos. One clear neighbor frame is enough for an efficient interpolation. Even when all neighbor frames are blurry, the partially clear areas of these frames can complement each other. We will analyze the efficiency of this interpolation in \Sref{sec:exp}.

\subsection{Combining enhanced and interpolated frames}

We use a simple network module to combine the output images from the enhancement and interpolation branches. The combination module is a simple architecture, containing two downsampling layers, followed by nine ResNet blocks and two upsampling layers. 

Early fusion is another approach to fuse the results from the enhancement and interpolation branches. Instead of fusing the final output images, we have also experimented with fusing the intermediate feature maps, five from the enhancement branch and two from the interpolation branch. But early fusion is not as good as late fusion, as will be seen in our ablation studies described below. 

\subsection{Training Loss}
For both interpolation and enhancement branches, we use $L_1$ loss between the predicted frame and ground truth. In addition, we also use spatial gradient loss~\cite{tran2018extreme} to enhance the sharpness of the reconstructed face. Let $G_x$ and $G_y$ be the gradient map respect to coordinate $x$ and $y$ respectively.
The training losses of the enhancement and interpolation branches are:
\begin{align}
    \mathcal{L}_{in} = &||I^{gt} - I^{in}||_1 + ||G_x(I^{gt}) - G_x(I^{in})||_1  \nonumber \\
                       & + ||G_y(I^{gt}) - G_y(I^{in})||_1,  \\
    \mathcal{L}_{en} = & ||I^{gt} - I^{en}||_1 + ||G_x(I^{gt}) - G_y(I^{en})||_1 \nonumber  \\
                       & + ||G_y(I^{gt}) - G_y(I^{en})||_1. 
\end{align}
where $I^{en}$ and $I^{in}$ are the image outputs of the enhancement and interpolation branches, and $I^{gt}$ is the ground truth frame. After training the enhancement and interpolation branches, we train the combination module to combine $I^{en}$ and $I^{in}$. For the combination module, we only use $L_1$ loss for training, i.e., $\mathcal{L}_{combine} = ||I^{gt} - \hat{I}||_1$, where $\hat{I}$ is the final output of our deblurring network. 

\section{Experiments}\label{sec:exp}

We compared FineNet with five state-of-the-art deblurring methods: two natural video deblurring methods~\cite{tao2018scale,wang2019edvr}, a natural image deblurring method~\cite{kupyn2018deblurgan}, a facial image deblurring method \cite{yasarla2019deblurring}, and a facial video deblurring method~\cite{Ren-ICCV-2019}. We used three datasets in our experiments, and all methods were trained on the same set of training data until convergence. We used PSNR and SSIM for quantiative evaluation and an user study for qualiative evaluation.


\subsection{Datasets}

We used the training videos of VoxCeleb for training the deblurring models, and we evaluated them on the test videos of VoxCeleb and 300VW datasets as well as  a self-collected dataset of blurry Youtube videos.

\setlength{\tabcolsep}{0pt}
\begin{figure*}[t]
\tiny
\begin{center}
\begin{tabular}{llllllll}
\multicolumn{1}{c}{Low quality} & 
\multicolumn{1}{c}{DeblurGAN \cite{kupyn2018deblurgan}} & 
\multicolumn{1}{c}{UMSN \cite{yasarla2019deblurring}} & 
\multicolumn{1}{c}{Ren et al. \cite{Ren-ICCV-2019}} & 
\multicolumn{1}{c}{EDVR \cite{wang2019edvr}} & 
\multicolumn{1}{c}{SRN \cite{tao2018scale}} & 
\multicolumn{1}{c}{Ours} &
\multicolumn{1}{c}{GT}\\
\cellimgsmall{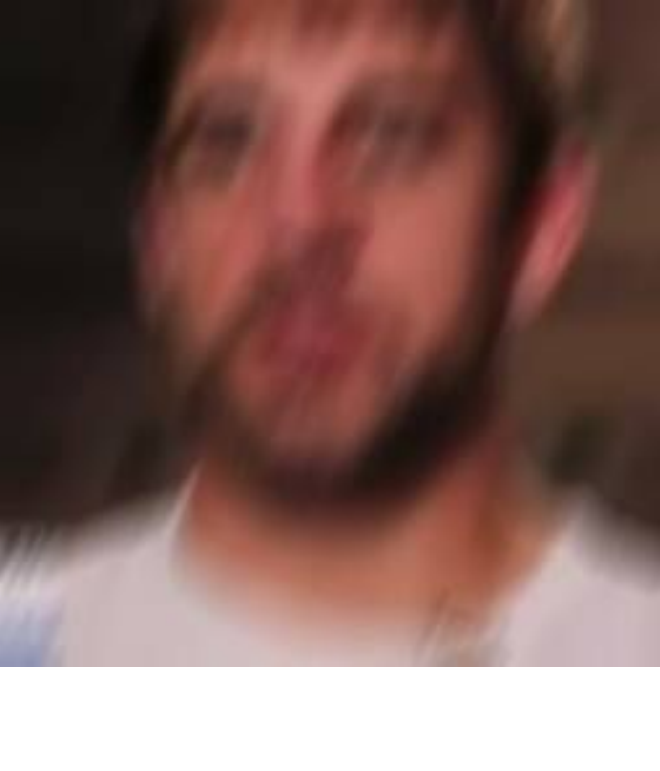} &
\cellimgsmall{images/quan1/face/deblurGAN.pdf} &
\cellimgsmall{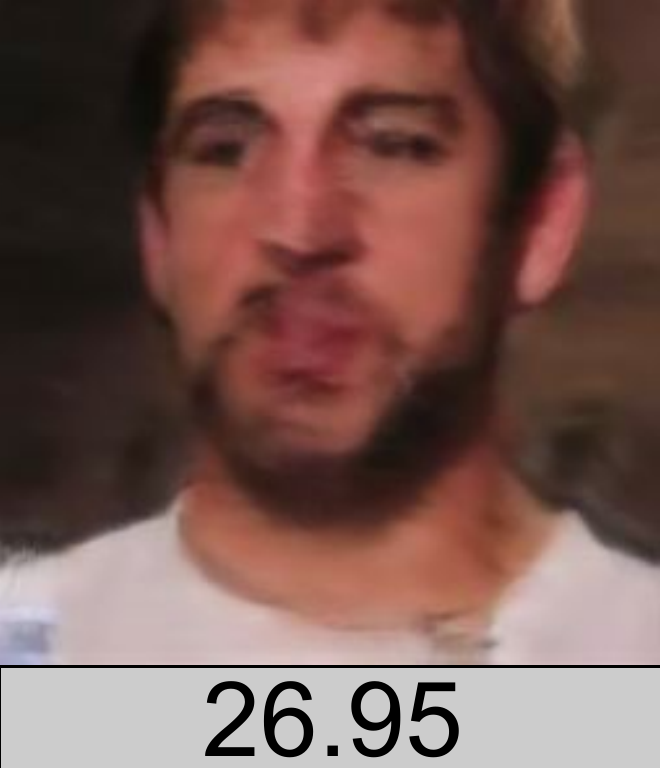} &
\cellimgsmall{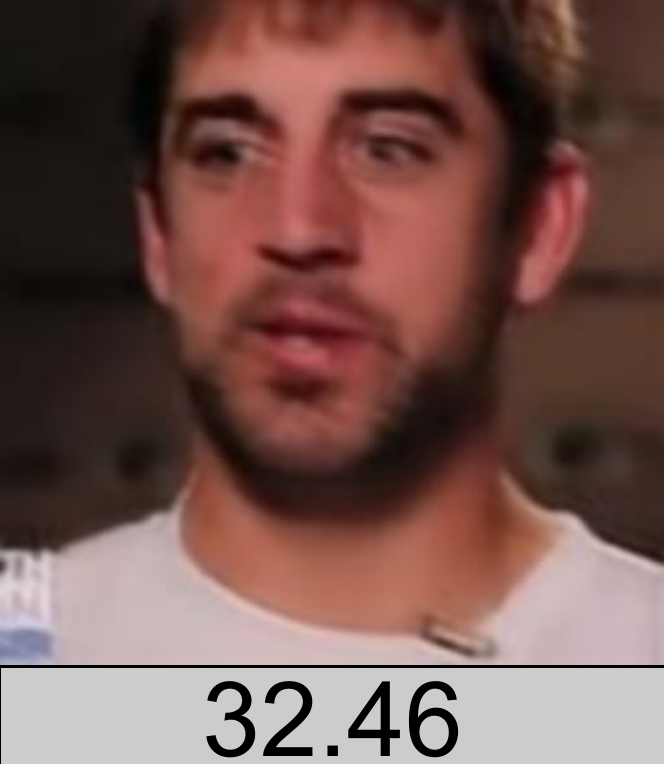} &
\cellimgsmall{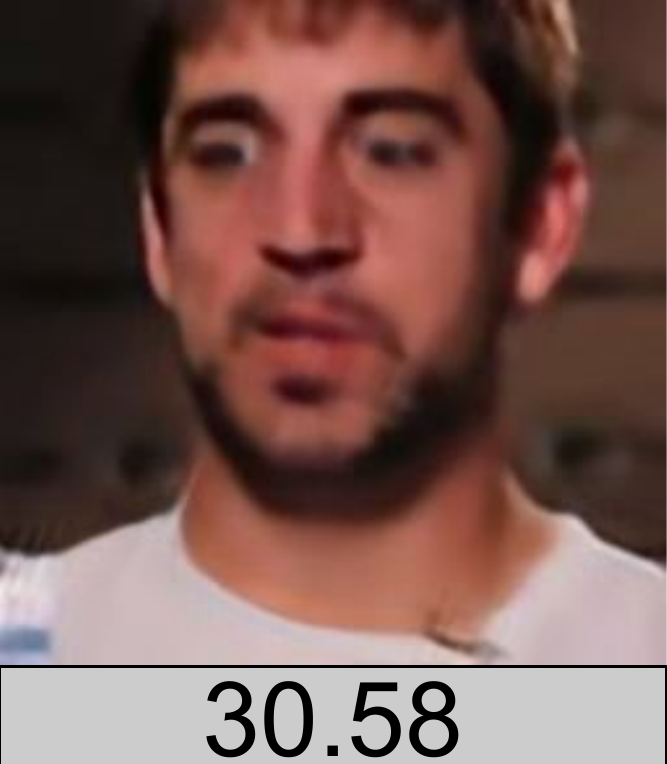} &
\cellimgsmall{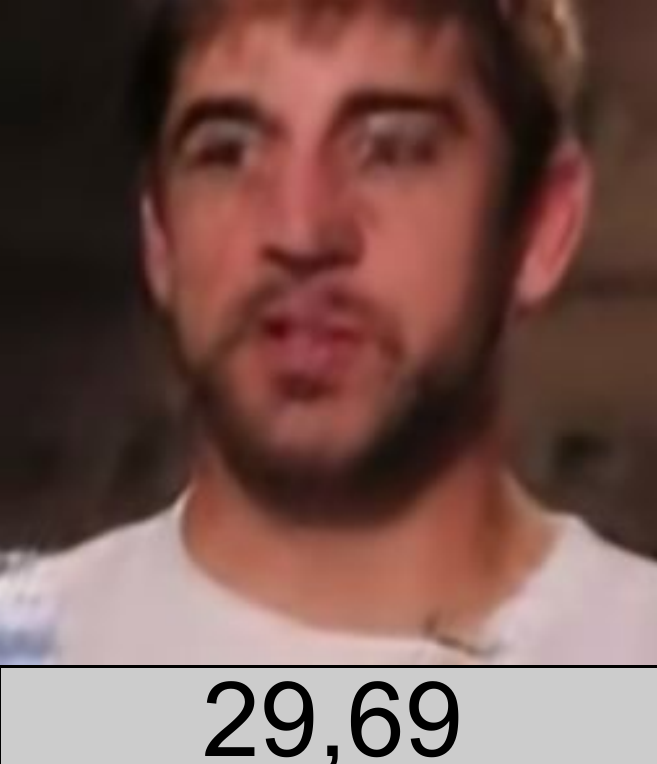} &
\cellimgsmall{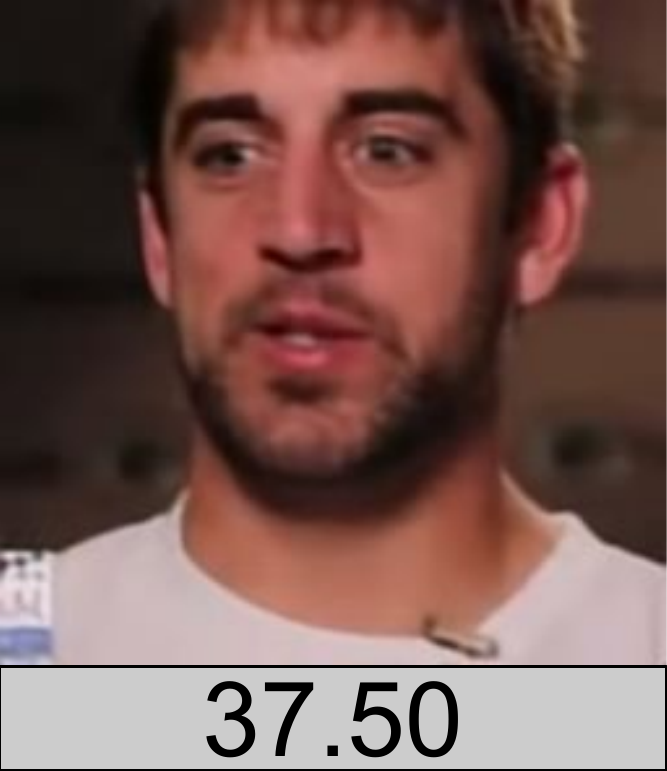} &
\cellimgsmall{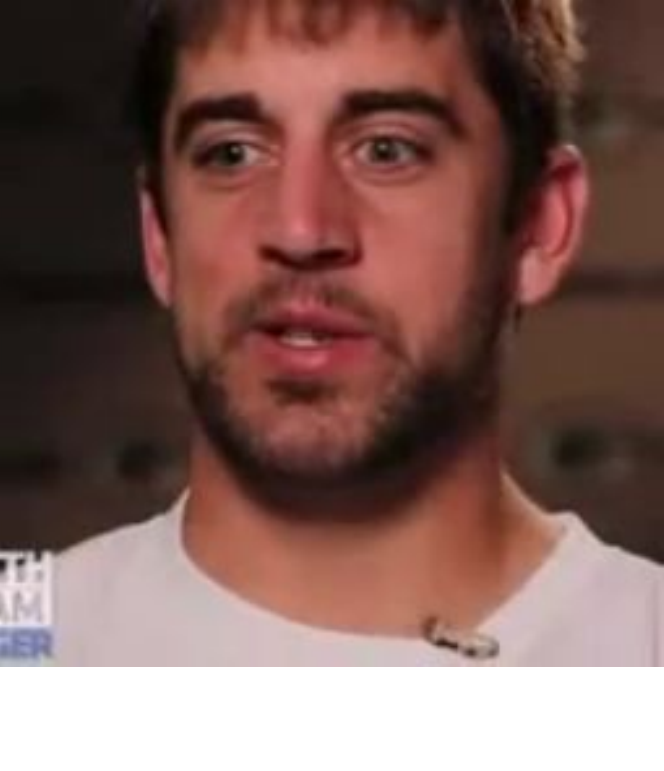}\\
\cellimgsmall{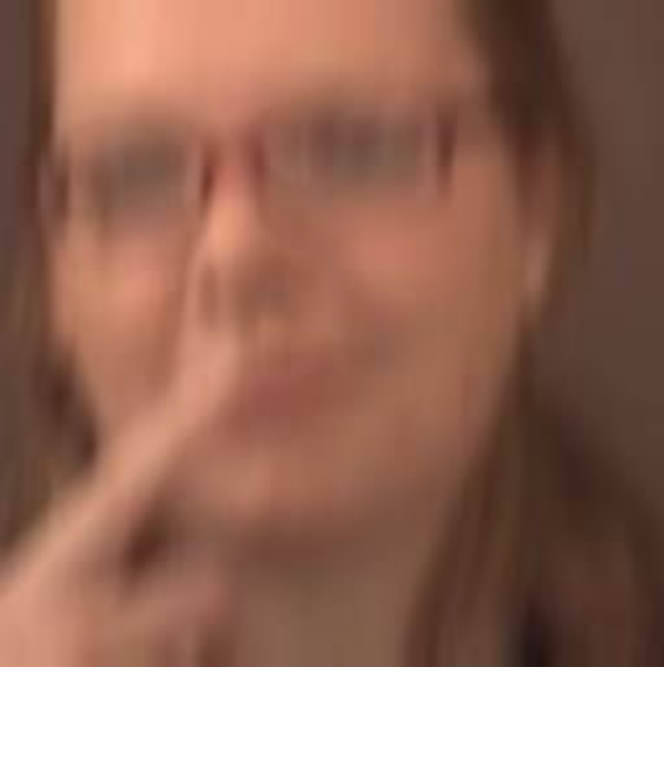} &
\cellimgsmall{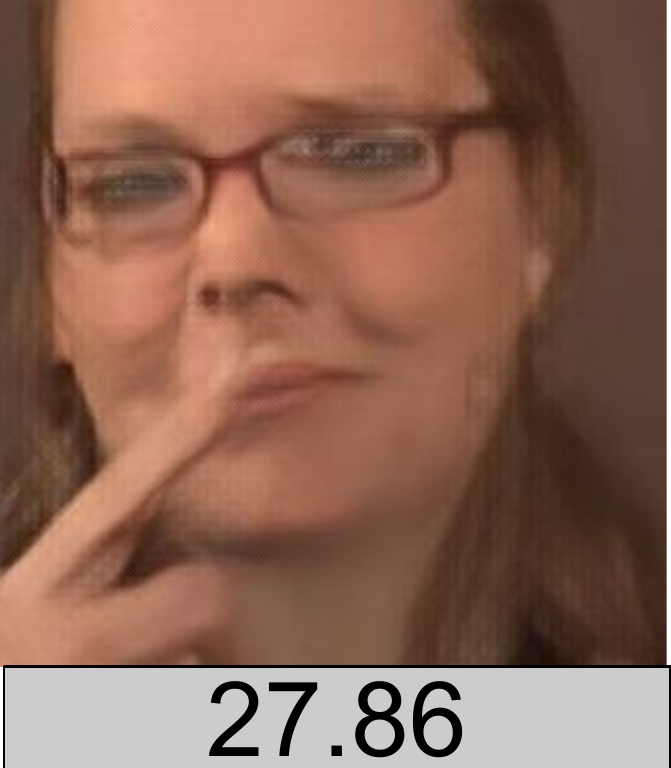} &
\cellimgsmall{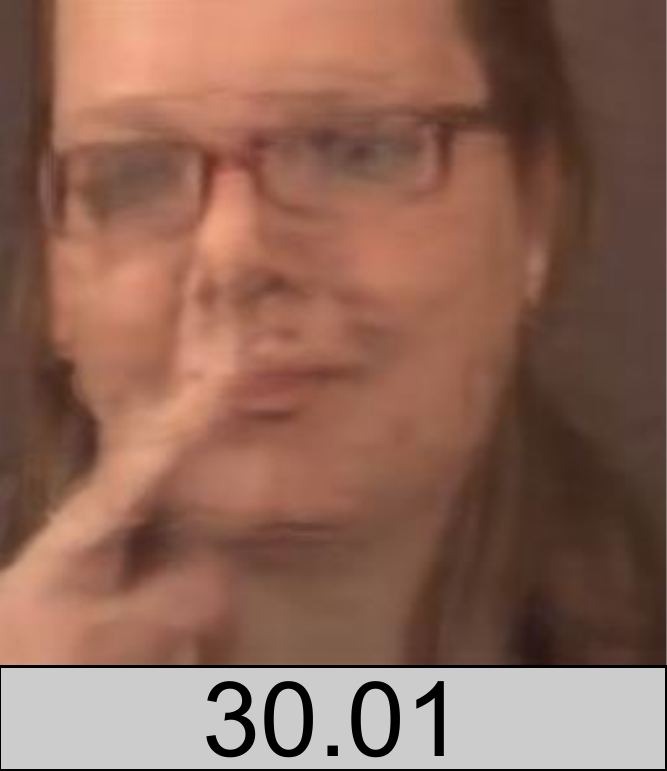} &
\cellimgsmall{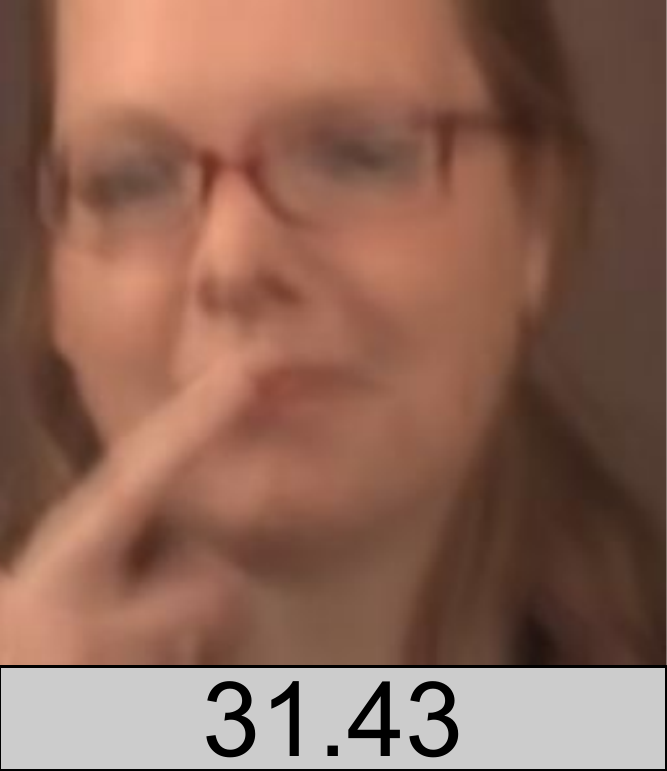} &
\cellimgsmall{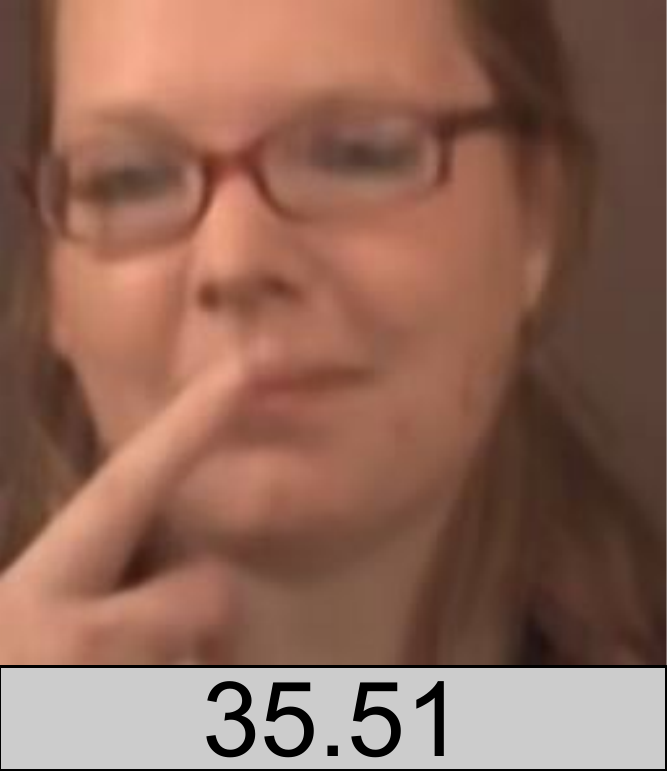} &
\cellimgsmall{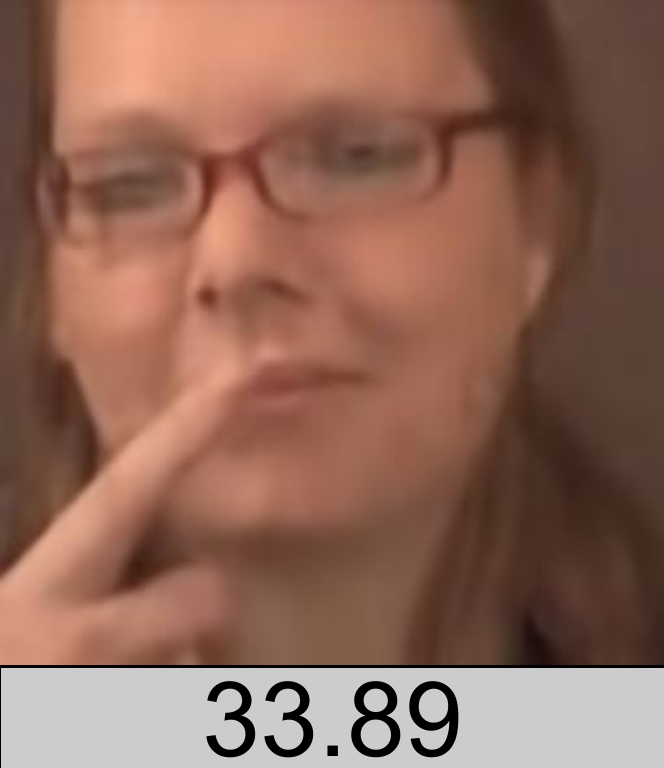} &
\cellimgsmall{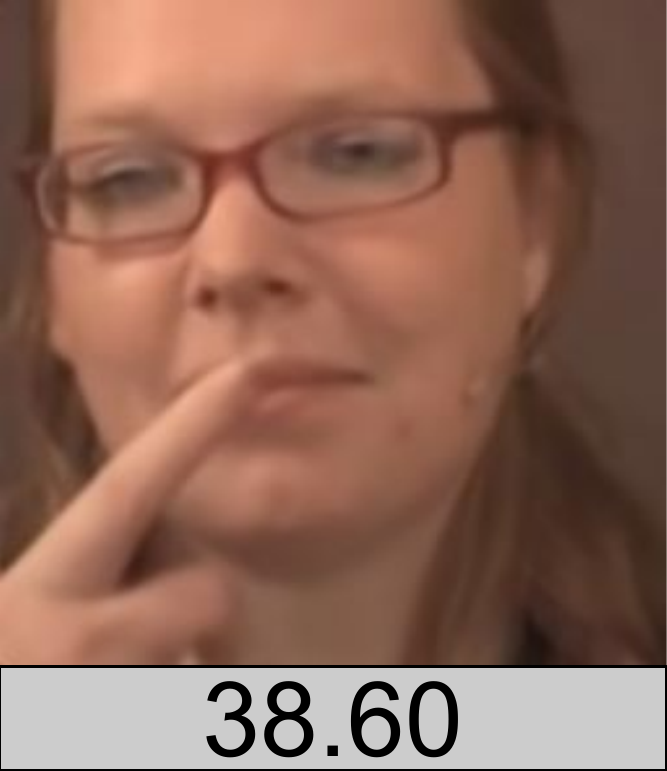} &
\cellimgsmall{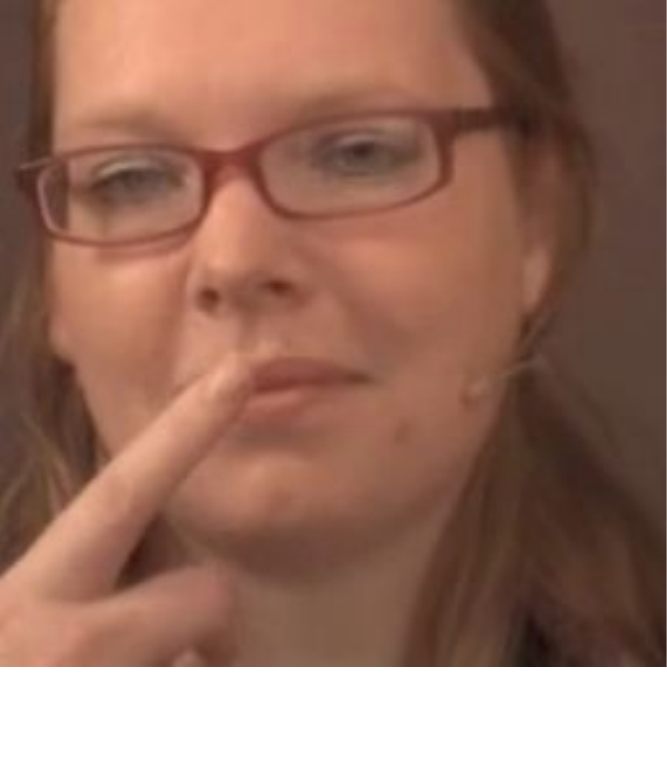}\\
\cellimgsmall{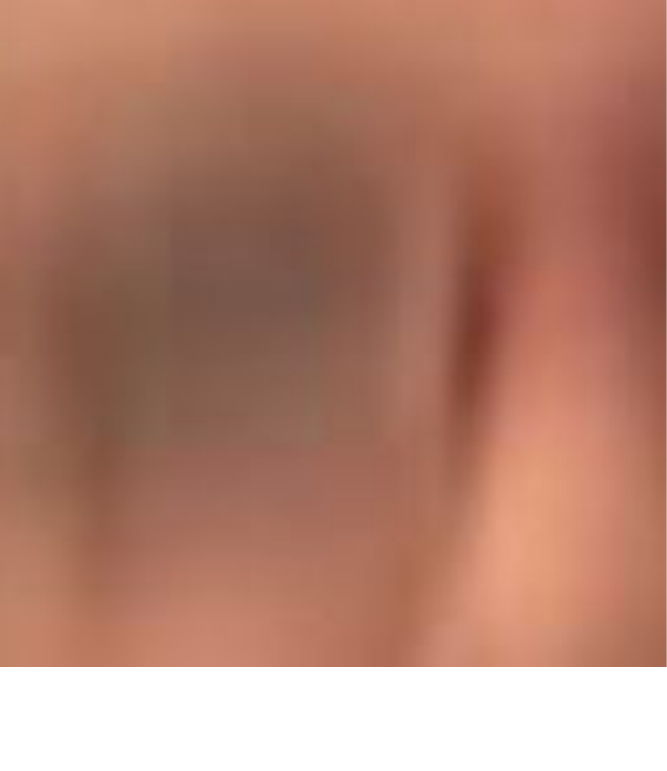} &
\cellimgsmall{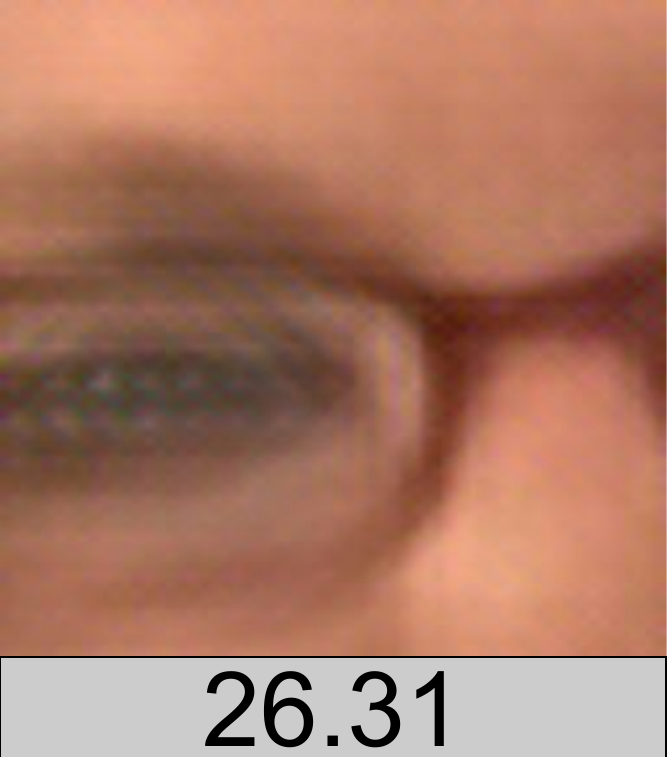} &
\cellimgsmall{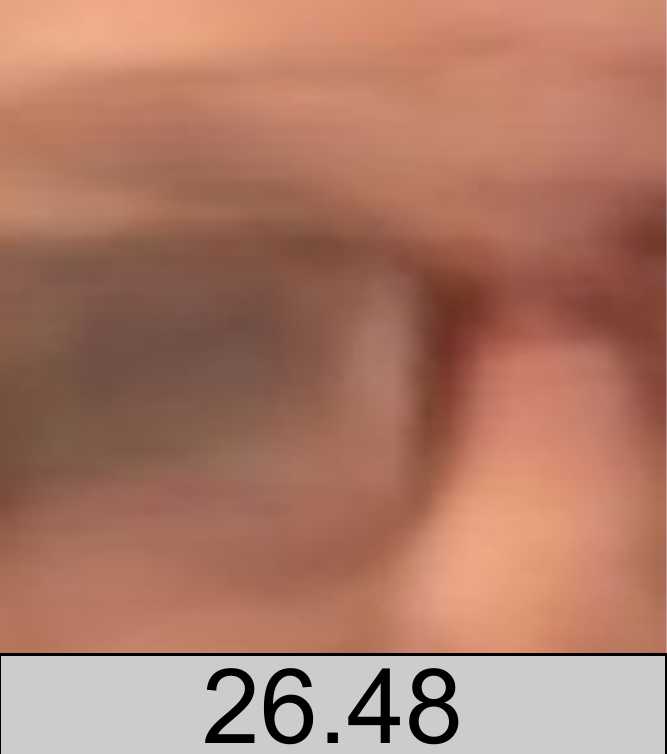} &
\cellimgsmall{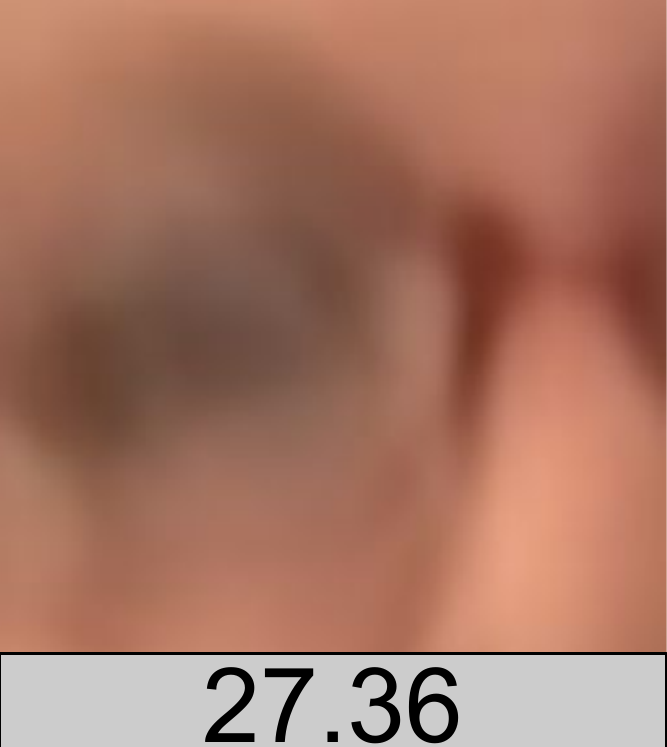} &
\cellimgsmall{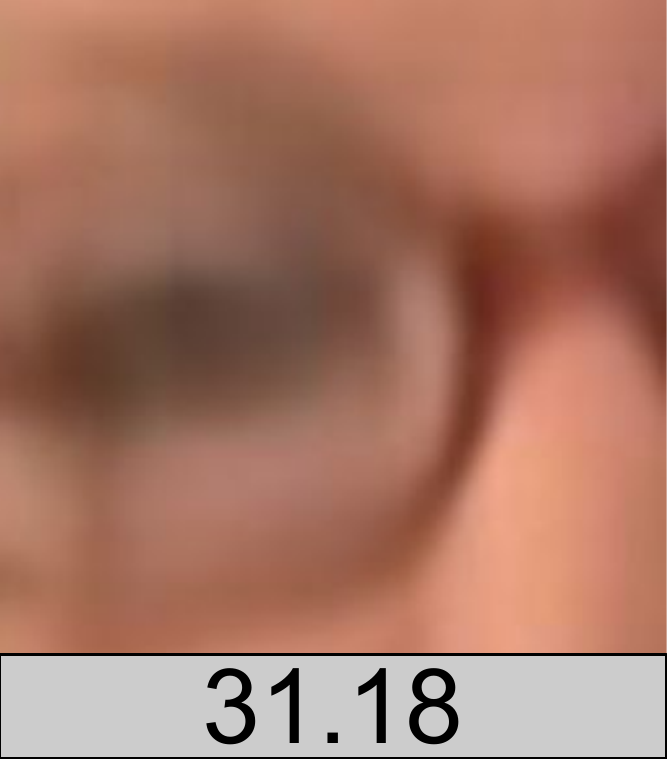} &
\cellimgsmall{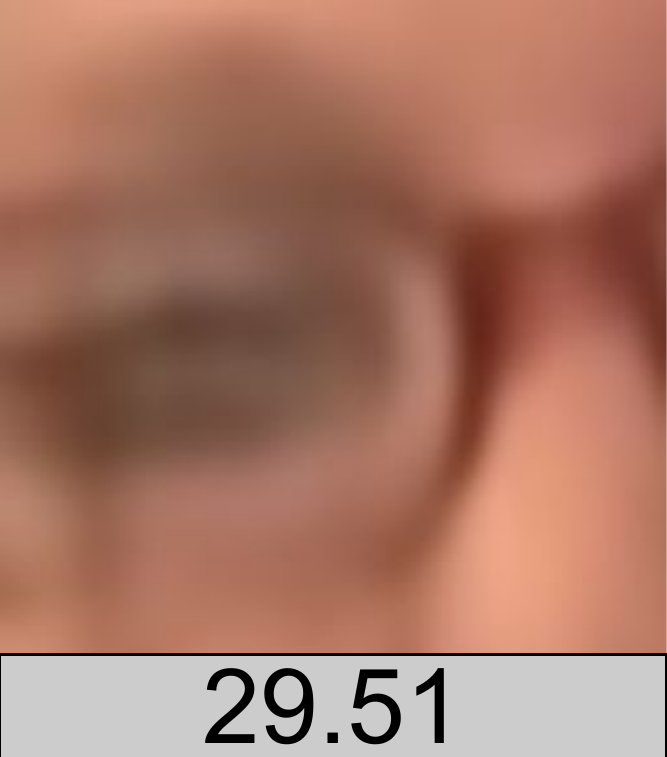} &
\cellimgsmall{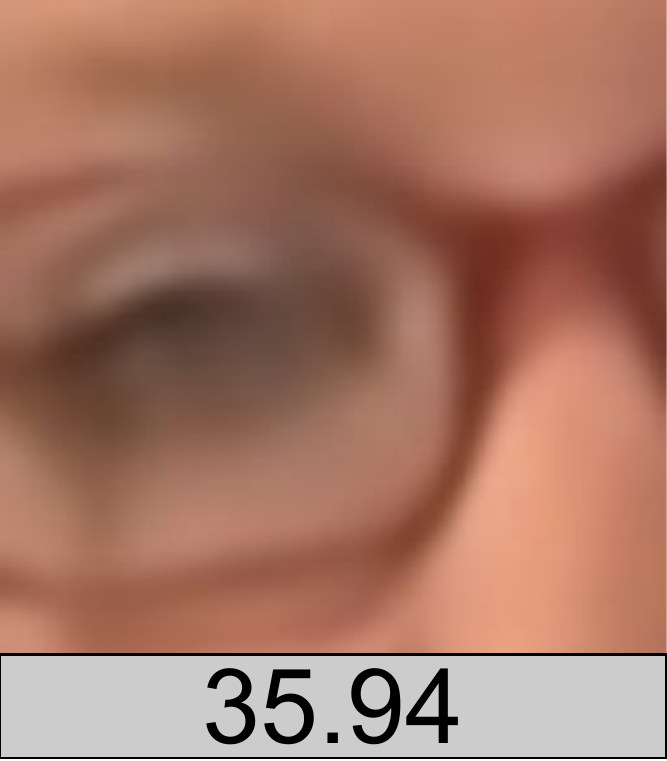} &
\cellimgsmall{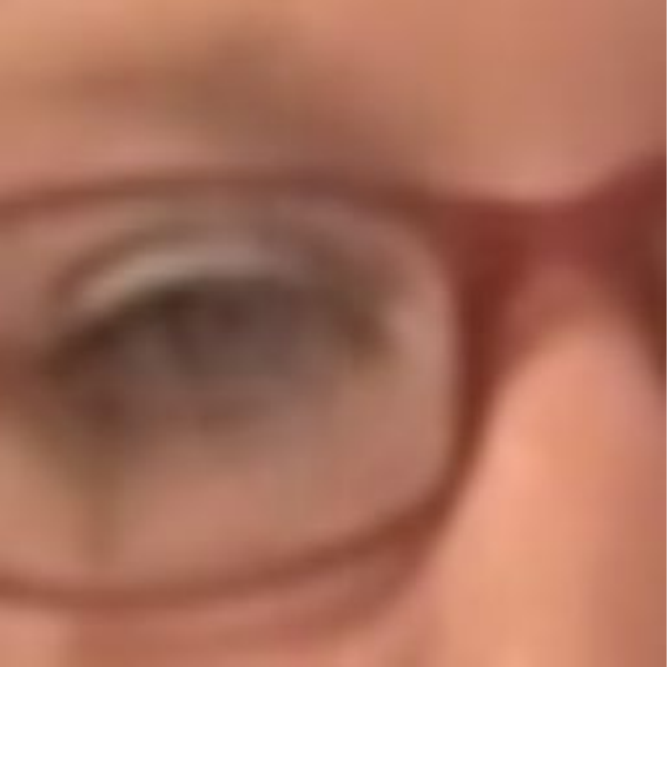}\\
\end{tabular}
\end{center}
\vskip -0.05in
\caption{Deblurring result on two examples from VoxCeleb dataset (top) and 300VW (bottom). Here we only show PSNR of the output image relative to the ground truth.
\vspace{-2mm}
\label{fig:quanex}}
\end{figure*}
\setlength{\tabcolsep}{1.4pt}

\myheading{VoxCeleb~\cite{Chung18b}.} This is a large dataset of face video clips collected from Youtube with about 146K training videos and 6K testing videos. Due to its large size, we randomly choose around 1300 videos from the training videos, approximately 700K frames, to create the training set. We use~\cite{imgaug} to synthesize blurred images. \citet{sun2019fab} suggested generating motion blur based on estimated face movement. However, constraining motion blur by a noisy and potentially inaccurate movement estimation limits the trained deblurring model's power. We empirically found that the models trained with random motion blur achieved higher performance. Therefore, we use random blur kernels to generate training data in our experiments. We chose 130 videos from the testing set for evaluation.

\myheading{300VW~\cite{shen2015first}.} This dataset contains high-quality and sharp face videos. We used 30 videos from this dataset for testing, evaluating the generalization ability of the trained models to a new domain. We did not use videos from this dataset for training the deblurring models. 

\myheading{Blurry Youtube.}
In addition to evaluating our method on synthetically blurred images, we also test it on 100 low-quality face videos from Youtube that contain blurry frames. Qualitative experiments show that our model provides clearer deblurring results than other methods.


\subsection{Implementation Details}
We train our network with mini-batch size of 12 images. We use Adam optimizer with $\beta_0=0.9$ and $\beta_1=0.999$. The initial learning rate is $10^{-4}$, and training converges after around 200,000 iterations.

%
Following \cite{wang2019edvr}, we use a two-stage strategy. In the first stage, we use a small deblurring module to remove noise and partially deblur the video. In the second stage, we will handle severely blurry frames. This strategy improves the performance of the alignment and interpolation modules.

\subsection{Comparison with State-of-the-art Methods}

\setlength{\tabcolsep}{3pt}
\begin{table}[t]
\begin{center}
\caption{Quantitative results on the VoxCeleb and 300VW testing sets. FineNet signficantly outperforms all other methods. \label{table:quanres}}
\label{table:headings}
\begin{tabular}{lrrrr}
\toprule 
& \multicolumn{2}{c}{VoxCeleb} & \multicolumn{2}{c}{300VW} \\
\cmidrule(lr){2-3}  \cmidrule(lr){4-5}
Method & PSNR & SSIM  & PSNR & SSIM \\
\midrule
DeblurGAN \cite{kupyn2018deblurgan}  & 26.45 & 0.4418 & 26.83 & 0.6228\\
UMSN \cite{yasarla2019deblurring} & 31.08 & 0.9035 & 31.35 & 0.9538\\
Ren et al. \cite{Ren-ICCV-2019} & 32.26 & 0.9089 & 33.05 & 0.9278\\
SRN-Deblur \cite{tao2018scale} & 34.71 & 0.9342 & 34.15 & 0.9501\\
EDVR \cite{wang2019edvr} & 34.77 & 0.8409 & 29.12 & 0.8775\\
\textbf{FineNet (proposed)} & \textbf{37.04} & \textbf{0.9612} & \textbf{38.22} & \textbf{0.9700}\\
\bottomrule 
\end{tabular}
\end{center}
\vspace{-5mm}
\end{table}
 
The average PSNR and SSIM values of all methods on the VoxCeleb and 300VW test sets are reported in \Tref{table:quanres}. As can be seen, FineNet outperforms other methods by a large margin. More specifically, on VoxCeleb, FineNet achieves PSNR of 37.04, while the PSNR of the second-best method is only 34.77. The performance gap between FineNet and the second-best one is even wider on the 300VW test set (c.f., 38.22 and 34.15). The models using face identity \cite{Ren-ICCV-2019,yasarla2019deblurring} fail to produce a clear output in case of severely blurred target frame. Meanwhile, FineNet still generates high-quality output with sharp component boundaries, even in that adverse condition, as shown in \Fref{fig:quanex}.

We notice a significant performance drop of the EDVR model on the cross-domain task, i.e., testing on 300VW data. As mentioned by~\citet{wang2019edvr}, the EDVR model is highly dataset-biased. Therefore, the performance would decrease when there are differences between the domains of the training and testing videos.

\setlength{\tabcolsep}{0pt}
\begin{figure*}[t]
\tiny
\begin{center}
\begin{tabular}{lllllll}
\cellimg{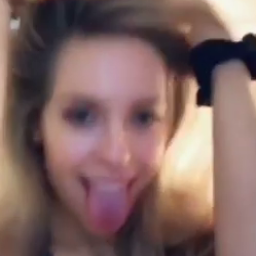} &
\cellimg{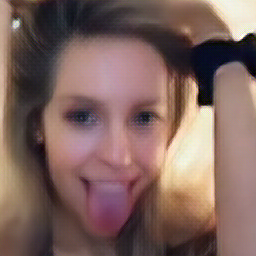} &
\cellimg{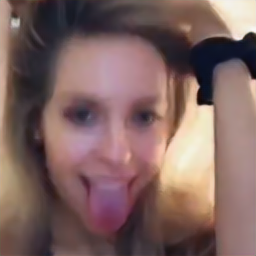} &
\cellimg{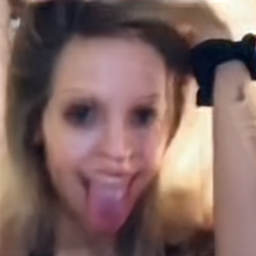} &
\cellimg{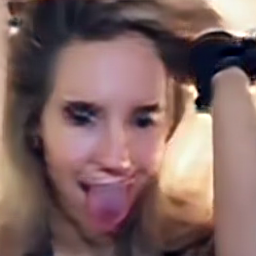} &
\cellimg{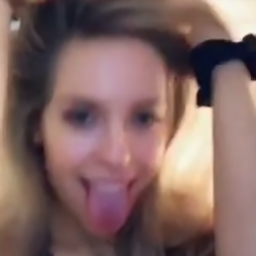} &
\cellimg{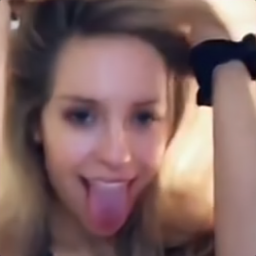}\\
\cellimg{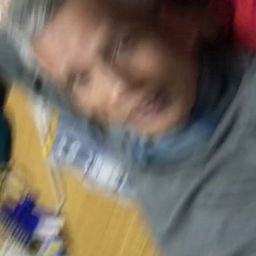} &
\cellimg{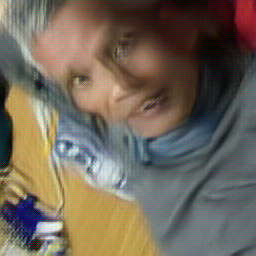} &
\cellimg{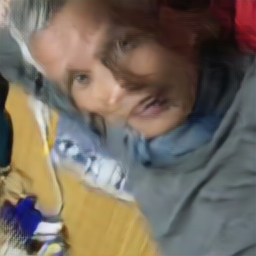} &
\cellimg{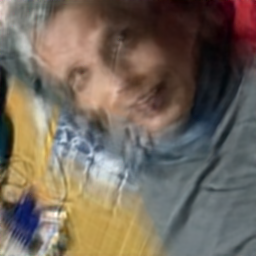} &
\cellimg{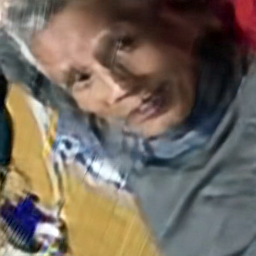} &
\cellimg{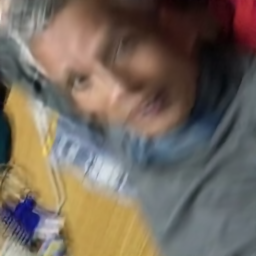} &
\cellimg{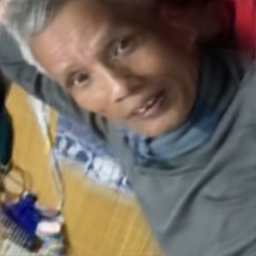}\\
\cellimg{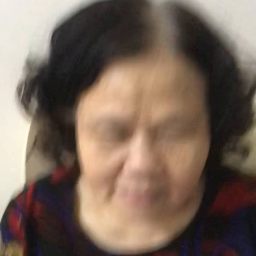} &
\cellimg{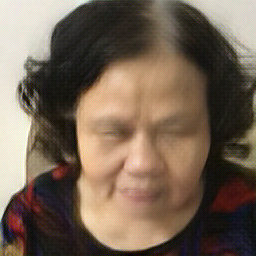} &
\cellimg{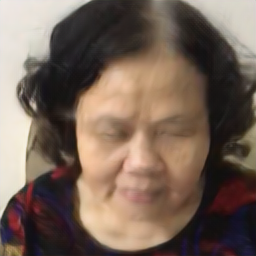} &
\cellimg{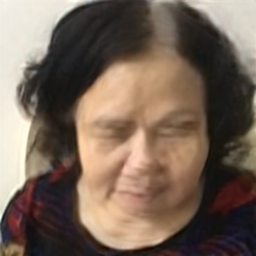} &
\cellimg{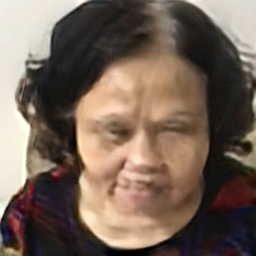} &
\cellimg{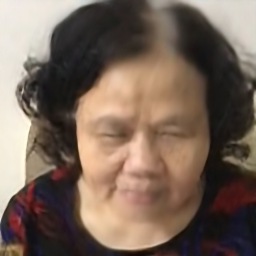} &
\cellimg{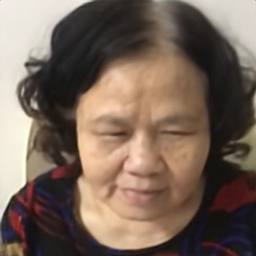}\\

\multicolumn{1}{c}{Low quality} & 
\multicolumn{1}{c}{DeblurGAN \cite{kupyn2018deblurgan}} & 
\multicolumn{1}{c}{UMSN \cite{yasarla2019deblurring}} & 
\multicolumn{1}{c}{Ren et al. \cite{Ren-ICCV-2019}} & 
\multicolumn{1}{c}{EDVR \cite{wang2019edvr}} & 
\multicolumn{1}{c}{SRN-Deblur \cite{tao2018scale}} & 
\multicolumn{1}{c}{Ours}\\
\end{tabular}
\end{center}
\vspace{-2mm}
\caption{Qualitative results on real blur video. We can easily see that the face components like eyes or nose that our model generated are sharper than other models. \label{qualres}}
\end{figure*}

Besides deblurring methods, we also compare FineNet with state-of-the-art interpolation methods, including \cite{bao2019depth} and \cite{sung2019generation}, on the 300VW dataset. Our method achieves 1.2db higher in terms of PSNR than the best method. Experiment details can be found in supplementary materials.

\subsection{Ablation Studies}\label{sec:ablation}

\setlength{\tabcolsep}{4pt}
\begin{table}[t]
\centering
\caption{Benefits of the FOC module and its landmark input on both Enhancement and Interpolation branches. The numbers are reported on 300VW dataset. \label{tab:FOG_ablation}}
\begin{tabular}{cccc}
\toprule
& & Enhancement & Interpolation\\
\cmidrule(lr){3-3} \cmidrule(lr){4-4}
FOC & Landmarks & PSNR/SSIM & PSNR/SSIM\\
\midrule
x & x & 29.12/0.8775 & 35.49/0.9490\\
\checkmark & x & 37.37/0.9590 & 36.36/0.9598 \\
\checkmark & \checkmark & 37.81/0.9649 & 36.87/0.9633\\

\bottomrule
\end{tabular}	
\vspace{-3mm}
\end{table}

We perform several ablation studies to understand the contribution of various components of FineNet. The first set of experiments are to test: (1) the importance of having the FOC modules for computing the location offsets for deformable convolution kernels; and (2) the importance of using landmark heatmaps in the FOC modules. \Tref{tab:FOG_ablation} reports the results on these experiments. Note that we propose to use FOC modules in both the enhancement and interpolation branches, so we actually perform two sets of ablation studies here. As can be observed, there is a clear benefit of using FOC modules; the PSNR drops sharply if we do not use FOC in the enhancement branch. The landmark heatmaps are also useful, for both enhancement and interpolation branches. Using the landmark heatmaps of two frames, FOC can predict the face movement and produce accurate offsets for deformable convolution.



\begin{figure}[t]
    \centering
\includegraphics[width=.5\textwidth]{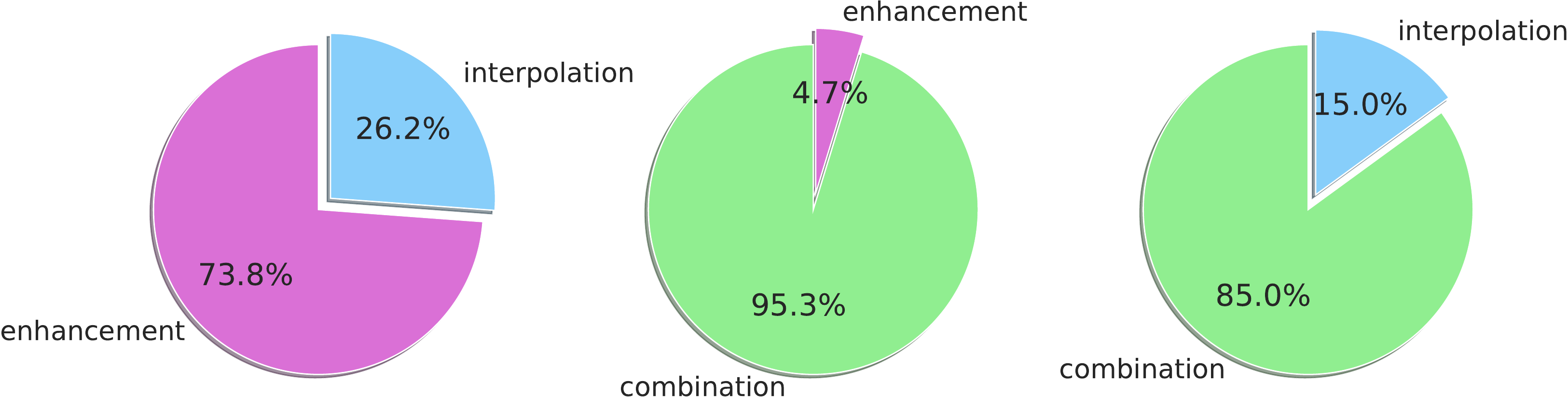}
\caption{Pairwise comparisons between the enhancement branch, the interpolation branch, and their combination on the 300VW dataset. For each pair of methods, the pie plot reports the number of times one method has a higher quantitative result than the other.}
\label{fig:pie}
\vspace{-3mm}
\end{figure}

Next, we evaluate the contribution of the enhancement and interpolation branches toward the final model. We report deblurring results on the 300VW dataset in \Tref{tab:Branch_ablation}. As can be seen, the best result is obtained when both branches are used. The need for using both branches has been justified with \Fref{fig:teaser}. On the one hand, the interpolation frame is useful when the blur is too severe for proper enhancement. On the other hand, if the target frame is already in good quality, it is easier to enhance it. FineNet combines the two methods to avoid the limitations of both. In \Fref{ablimg}, the first row is the case in which the output of the interpolation branch is better than the enhancement branch, while in the second row, the interpolation branch fails to generate sharp facial components. The combination of both branches yields the best results in both cases. As shown in the second and the third plots of \Fref{fig:pie}, the combination method outperforms enhancement and interpolation for 95.3\% and 85.0\% of time, respectively.



To verify our fusion scheme, we also report the performance of the early fusion approach in the third row of \Tref{tab:Branch_ablation}. Unlike our late fusion, it gives only a minor gain. This is because the feature maps from the enhancement branch outnumber the feature maps o the interpolated branch, making early fusion ineffective.

While landmark input is helpful in overall, landmarks estimated from severely blurry frames are unreliable and may hurt the system performance. To verify that, we test FineNet with ``oracle'' landmarks computed from the ground-truth frames and report in the last row of \Tref{tab:Branch_ablation}. There is no gain in PSNR, indicating the robustness of our system with respect to landmark localization error. This is because we only use landmark heatmaps as an implicit prior for calculating the deformable kernel offsets. This approach is robust to the failure of the landmark detector, unlike the methods that explicitly use facial structure priors.


\setlength{\tabcolsep}{3pt}
\begin{table}[t]
\centering
\caption{Benefits of Enhancement-Interpolation fusion. The numbers are reported on 300VW dataset. Enh and Int denote Enhancement and Interpolation branches respectively \label{tab:Branch_ablation}}
\begin{tabular}{ccccc}
\toprule
Enh & Int & Late fusion & Oracle landmarks & PSNR/SSIM \\
\midrule 
\checkmark & x &  & & 37.81/0.9649\\
x & \checkmark &  & & 36.87/0.9633\\
\checkmark & \checkmark  & x & & 37.88/0.9655\\
\checkmark & \checkmark & \checkmark & & 38.22/0.9700\\
\midrule 
\checkmark & \checkmark & \checkmark & \checkmark & 38.22/0.9698 \\

\bottomrule
\vspace{-4mm}
\end{tabular}
\end{table}

For the interpolation module, we also experimented with using four frames for both forward and backward interpolation. However, there is no significant difference in performance for using either three or four frames. The PSNR/SSIM of the two methods are 36.83/0.9632 and 36.87/0.9633, respectively.

\newcommand{\cellimgtwo}[1]{
    \includegraphics[width=1.6cm, height=1.6cm]{#1}
}
\setlength{\tabcolsep}{0pt}
\begin{figure}[t]
\begin{center}
\begin{tabular}{lllll}
\multicolumn{1}{c}{\small Low quality} &
\multicolumn{1}{c}{\small EDVR \cite{wang2019edvr}} & 
\multicolumn{1}{c}{\small Interpolation} &
\multicolumn{1}{c}{\small Enhancement} &
\multicolumn{1}{c}{\small Combination} \\
\cellimgtwo{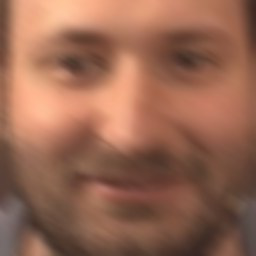} &
\cellimgtwo{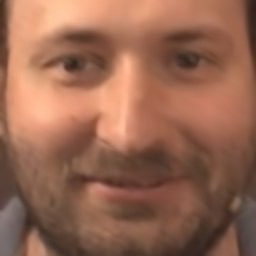} &
\cellimgtwo{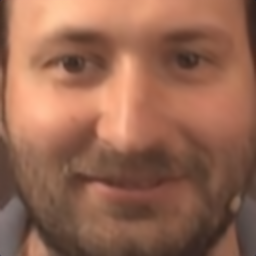} &
\cellimgtwo{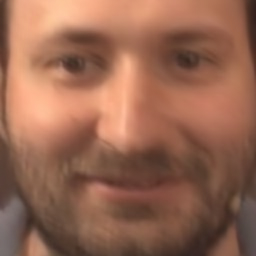} &
\cellimgtwo{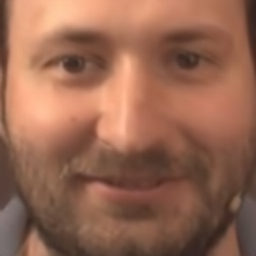}\\
 & \multicolumn{1}{c}{37.44} & 
\multicolumn{1}{c}{38.54} &
\multicolumn{1}{c}{38.12} &
\multicolumn{1}{c}{38.83}\\
\cellimgtwo{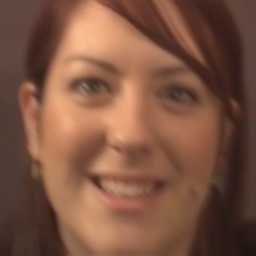} &
\cellimgtwo{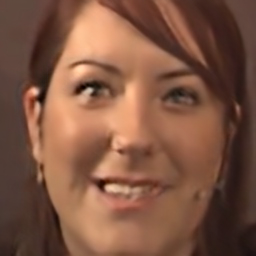} &
\cellimgtwo{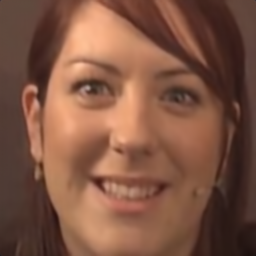} &
\cellimgtwo{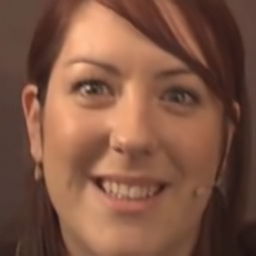} &
\cellimgtwo{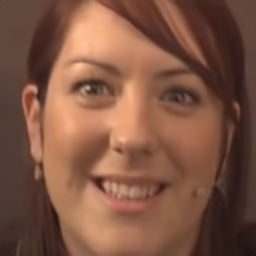}\\
 & \multicolumn{1}{c}{27.71} & 
\multicolumn{1}{c}{39.09} &
\multicolumn{1}{c}{40.01} &
\multicolumn{1}{c}{40.33}\\
\end{tabular}
\end{center}
\vspace{-3mm}
\caption{Result of each module on two examples in 300VW testing set. We use PSNR to evaluate the quality of images. \label{ablimg}}
\vspace{-3mm}
\end{figure}
\setlength{\tabcolsep}{1.4pt}

\subsection{Evaluation on real blur videos}
Finally, to validate the practical usefulness of FineNet, we run it and other state-of-the-art methods on the real low-quality Youtube videos. The results are shown in \Fref{qualres}. Those videos contain real blur caused by camera shakes, rapid face movement, or both. The frames generated by our model look better than the outputs of other methods. We conducted an user study where 46 users was asked to select the best deblurred image among the results of FineNet and other methods for 23 test cases. Impressively,  FineNet was selected 91.3\% of the time. 



\section{Conclusions}
\vspace{-0.2cm}
We have presented FineNet, a network for video deblurring. FineNet contains two parallel branches for frame enhancement and frame interpolation. We have also introduced a novel module for calculating the spatial offsets between two facial feature maps, using landmark heatmaps. Experiments on a number of real and synthetic blurry frames demonstrate the good performance of our method.


\begin{thebibliography}{31}
\providecommand{\natexlab}[1]{#1}
\providecommand{\url}[1]{\texttt{#1}}
\expandafter\ifx\csname urlstyle\endcsname\relax
  \providecommand{\doi}[1]{doi: #1}\else
  \providecommand{\doi}{doi: \begingroup \urlstyle{rm}\Url}\fi

\bibitem[Bao et~al.(2019)Bao, Lai, Ma, Zhang, Gao, and Yang]{bao2019depth}
Wenbo Bao, Wei-Sheng Lai, Chao Ma, Xiaoyun Zhang, Zhiyong Gao, and Ming-Hsuan
  Yang.
\newblock Depth-aware video frame interpolation.
\newblock In \emph{Proceedings of the IEEE Conference on Computer Vision and
  Pattern Recognition}, pages 3703--3712, 2019.

\bibitem[Bulat and Tzimiropoulos(2017{\natexlab{a}})]{bulat2017binarized}
Adrian Bulat and Georgios Tzimiropoulos.
\newblock Binarized convolutional landmark localizers for human pose estimation
  and face alignment with limited resources.
\newblock In \emph{Proceedings of the IEEE International Conference on Computer
  Vision}, pages 3706--3714, 2017{\natexlab{a}}.

\bibitem[Bulat and Tzimiropoulos(2017{\natexlab{b}})]{bulat2017far}
Adrian Bulat and Georgios Tzimiropoulos.
\newblock How far are we from solving the 2d \& 3d face alignment problem?(and
  a dataset of 230,000 3d facial landmarks).
\newblock In \emph{Proceedings of the IEEE International Conference on Computer
  Vision}, pages 1021--1030, 2017{\natexlab{b}}.

\bibitem[Caballero et~al.(2017)Caballero, Ledig, Aitken, Acosta, Totz, Wang,
  and Shi]{caballero2017real}
Jose Caballero, Christian Ledig, Andrew Aitken, Alejandro Acosta, Johannes
  Totz, Zehan Wang, and Wenzhe Shi.
\newblock Real-time video super-resolution with spatio-temporal networks and
  motion compensation.
\newblock In \emph{Proceedings of the IEEE Conference on Computer Vision and
  Pattern Recognition}, pages 4778--4787, 2017.

\bibitem[Chan and Wong(1998)]{chan1998total}
Tony~F Chan and Chiu-Kwong Wong.
\newblock Total variation blind deconvolution.
\newblock \emph{IEEE transactions on Image Processing}, 7\penalty0
  (3):\penalty0 370--375, 1998.

\bibitem[Chung et~al.(2018)Chung, Nagrani, and Zisserman]{Chung18b}
J.~S. Chung, A.~Nagrani, and A.~Zisserman.
\newblock Voxceleb2: Deep speaker recognition.
\newblock In \emph{INTERSPEECH}, 2018.

\bibitem[Dai et~al.(2017)Dai, Qi, Xiong, Li, Zhang, Hu, and
  Wei]{dai2017deformable}
Jifeng Dai, Haozhi Qi, Yuwen Xiong, Yi~Li, Guodong Zhang, Han Hu, and Yichen
  Wei.
\newblock Deformable convolutional networks.
\newblock In \emph{Proceedings of the IEEE international conference on computer
  vision}, pages 764--773, 2017.

\bibitem[Dong et~al.(2018)Dong, Yan, Ouyang, and Yang]{dong2018style}
Xuanyi Dong, Yan Yan, Wanli Ouyang, and Yi~Yang.
\newblock Style aggregated network for facial landmark detection.
\newblock In \emph{Proceedings of the IEEE Conference on Computer Vision and
  Pattern Recognition}, pages 379--388, 2018.

\bibitem[Jo et~al.(2018)Jo, Wug~Oh, Kang, and Joo~Kim]{jo2018deep}
Younghyun Jo, Seoung Wug~Oh, Jaeyeon Kang, and Seon Joo~Kim.
\newblock Deep video super-resolution network using dynamic upsampling filters
  without explicit motion compensation.
\newblock In \emph{Proceedings of the IEEE conference on computer vision and
  pattern recognition}, pages 3224--3232, 2018.

\bibitem[Jung et~al.(2020)Jung, Wada, Crall, Tanaka, Graving, Reinders, Yadav,
  Banerjee, Vecsei, Kraft, Rui, Borovec, Vallentin, Zhydenko, Pfeiffer, Cook,
  Fernández, De~Rainville, Weng, Ayala-Acevedo, Meudec, Laporte,
  et~al.]{imgaug}
Alexander~B. Jung, Kentaro Wada, Jon Crall, Satoshi Tanaka, Jake Graving,
  Christoph Reinders, Sarthak Yadav, Joy Banerjee, Gábor Vecsei, Adam Kraft,
  Zheng Rui, Jirka Borovec, Christian Vallentin, Semen Zhydenko, Kilian
  Pfeiffer, Ben Cook, Ismael Fernández, François-Michel De~Rainville,
  Chi-Hung Weng, Abner Ayala-Acevedo, Raphael Meudec, Matias Laporte, et~al.
\newblock {imgaug}.
\newblock \url{https://github.com/aleju/imgaug}, 2020.
\newblock Online; accessed 01-Feb-2020.

\bibitem[Kim et~al.(2018)Kim, Sajjadi, Hirsch, and
  Sch{\"o}lkopf]{kim2018spatio}
Tae~Hyun Kim, Mehdi~SM Sajjadi, Michael Hirsch, and Bernhard Sch{\"o}lkopf.
\newblock Spatio-temporal transformer network for video restoration.
\newblock In \emph{European Conference on Computer Vision}, pages 111--127.
  Springer, 2018.

\bibitem[Kupyn et~al.(2018)Kupyn, Budzan, Mykhailych, Mishkin, and
  Matas]{kupyn2018deblurgan}
Orest Kupyn, Volodymyr Budzan, Mykola Mykhailych, Dmytro Mishkin, and
  Ji{\v{r}}{\'\i} Matas.
\newblock Deblurgan: Blind motion deblurring using conditional adversarial
  networks.
\newblock In \emph{Proceedings of the IEEE conference on computer vision and
  pattern recognition}, pages 8183--8192, 2018.

\bibitem[Liu et~al.(2014)Liu, Chang, and Ma]{liu2014blind}
Guangcan Liu, Shiyu Chang, and Yi~Ma.
\newblock Blind image deblurring using spectral properties of convolution
  operators.
\newblock \emph{IEEE Transactions on image processing}, 23\penalty0
  (12):\penalty0 5047--5056, 2014.

\bibitem[Pan et~al.(2016)Pan, Sun, Pfister, and Yang]{pan2016blind}
Jinshan Pan, Deqing Sun, Hanspeter Pfister, and Ming-Hsuan Yang.
\newblock Blind image deblurring using dark channel prior.
\newblock In \emph{Proceedings of the IEEE Conference on Computer Vision and
  Pattern Recognition}, pages 1628--1636, 2016.

\bibitem[Ren et~al.(2019)Ren, Yang, Deng, Wipf, Cao, and Tong]{Ren-ICCV-2019}
Wenqi Ren, Jiaolong Yang, Senyou Deng, David Wipf, Xiaochun Cao, and Xin Tong.
\newblock Face video deblurring via 3d facial priors.
\newblock In \emph{IEEE International Conference on Computer Vision}, 2019.

\bibitem[Shen et~al.(2015)Shen, Zafeiriou, Chrysos, Kossaifi, Tzimiropoulos,
  and Pantic]{shen2015first}
Jie Shen, Stefanos Zafeiriou, Grigoris~G Chrysos, Jean Kossaifi, Georgios
  Tzimiropoulos, and Maja Pantic.
\newblock The first facial landmark tracking in-the-wild challenge: Benchmark
  and results.
\newblock In \emph{Proceedings of the IEEE international conference on computer
  vision workshops}, pages 50--58, 2015.

\bibitem[Shen et~al.(2018)Shen, Lai, Xu, Kautz, and Yang]{shen2018deep}
Ziyi Shen, Wei-Sheng Lai, Tingfa Xu, Jan Kautz, and Ming-Hsuan Yang.
\newblock Deep semantic face deblurring.
\newblock In \emph{Proceedings of the IEEE Conference on Computer Vision and
  Pattern Recognition}, pages 8260--8269, 2018.

\bibitem[Sun et~al.(2019)Sun, Wu, Liu, Yang, Wang, Zhou, Ye, and
  Qian]{sun2019fab}
Keqiang Sun, Wayne Wu, Tinghao Liu, Shuo Yang, Quan Wang, Qiang Zhou, Zuochang
  Ye, and Chen Qian.
\newblock Fab: A robust facial landmark detection framework for motion-blurred
  videos.
\newblock In \emph{Proceedings of the IEEE International Conference on Computer
  Vision}, pages 5462--5471, 2019.

\bibitem[Sung et~al.(2019)Sung, Shin, Kim, Lee, Park, Moon, and
  Kim]{sung2019generation}
Suk-Kyung Sung, Seungheon Shin, TaeYoung Kim, Jin-Yi Lee, Eunsu Park, Yong-Jae
  Moon, and Il-Hoon Kim.
\newblock Generation of high cadence sdo/aia images using a video frame
  interpolation method, superslomo.
\newblock \emph{The Bulletin of The Korean Astronomical Society}, 44\penalty0
  (2):\penalty0 44--1, 2019.

\bibitem[Tang et~al.(2018)Tang, Peng, Geng, Wu, Zhang, and
  Metaxas]{tang2018quantized}
Zhiqiang Tang, Xi~Peng, Shijie Geng, Lingfei Wu, Shaoting Zhang, and Dimitris
  Metaxas.
\newblock Quantized densely connected u-nets for efficient landmark
  localization.
\newblock In \emph{Proceedings of the European Conference on Computer Vision
  (ECCV)}, pages 339--354, 2018.

\bibitem[Tao et~al.(2017)Tao, Gao, Liao, Wang, and Jia]{tao2017detail}
Xin Tao, Hongyun Gao, Renjie Liao, Jue Wang, and Jiaya Jia.
\newblock Detail-revealing deep video super-resolution.
\newblock In \emph{Proceedings of the IEEE International Conference on Computer
  Vision}, pages 4472--4480, 2017.

\bibitem[Tao et~al.(2018)Tao, Gao, Shen, Wang, and Jia]{tao2018scale}
Xin Tao, Hongyun Gao, Xiaoyong Shen, Jue Wang, and Jiaya Jia.
\newblock Scale-recurrent network for deep image deblurring.
\newblock In \emph{Proceedings of the IEEE Conference on Computer Vision and
  Pattern Recognition}, pages 8174--8182, 2018.

\bibitem[Tian et~al.(2018)Tian, Zhang, Fu, and Xu]{tian2018tdan}
Yapeng Tian, Yulun Zhang, Yun Fu, and Chenliang Xu.
\newblock Tdan: Temporally deformable alignment network for video
  super-resolution.
\newblock \emph{arXiv preprint arXiv:1812.02898}, 2018.

\bibitem[Tran et~al.(2018)Tran, Hassner, Masi, Paz, Nirkin, and
  Medioni]{tran2018extreme}
Anh~Tuan Tran, Tal Hassner, Iacopo Masi, Eran Paz, Yuval Nirkin, and
  G{\'e}rard~G Medioni.
\newblock Extreme 3d face reconstruction: Seeing through occlusions.
\newblock In \emph{CVPR}, pages 3935--3944, 2018.

\bibitem[Valle et~al.(2019)Valle, Buenaposada, Vald{\'e}s, and
  Baumela]{valle2019face}
Roberto Valle, Jos{\'e}~M Buenaposada, Antonio Vald{\'e}s, and Luis Baumela.
\newblock Face alignment using a 3d deeply-initialized ensemble of regression
  trees.
\newblock \emph{Computer Vision and Image Understanding}, 189:\penalty0 102846,
  2019.

\bibitem[Wang et~al.(2019)Wang, Chan, Yu, Dong, and Change~Loy]{wang2019edvr}
Xintao Wang, Kelvin~CK Chan, Ke~Yu, Chao Dong, and Chen Change~Loy.
\newblock Edvr: Video restoration with enhanced deformable convolutional
  networks.
\newblock In \emph{Proceedings of the IEEE Conference on Computer Vision and
  Pattern Recognition Workshops}, pages 0--0, 2019.

\bibitem[Xue et~al.(2019)Xue, Chen, Wu, Wei, and Freeman]{xue2019video}
Tianfan Xue, Baian Chen, Jiajun Wu, Donglai Wei, and William~T Freeman.
\newblock Video enhancement with task-oriented flow.
\newblock \emph{International Journal of Computer Vision}, 127\penalty0
  (8):\penalty0 1106--1125, 2019.

\bibitem[Yang et~al.(2017)Yang, Liu, and Zhang]{yang2017stacked}
Jing Yang, Qingshan Liu, and Kaihua Zhang.
\newblock Stacked hourglass network for robust facial landmark localisation.
\newblock In \emph{Proceedings of the IEEE Conference on Computer Vision and
  Pattern Recognition Workshops}, pages 79--87, 2017.

\bibitem[Yasarla et~al.(2019)Yasarla, Perazzi, and
  Patel]{yasarla2019deblurring}
Rajeev Yasarla, Federico Perazzi, and Vishal~M Patel.
\newblock Deblurring face images using uncertainty guided multi-stream semantic
  networks.
\newblock \emph{arXiv preprint arXiv:1907.13106}, 2019.

\bibitem[Zhu et~al.(2019)Zhu, Hu, Lin, and Dai]{zhu2019deformable}
Xizhou Zhu, Han Hu, Stephen Lin, and Jifeng Dai.
\newblock Deformable convnets v2: More deformable, better results.
\newblock In \emph{Proceedings of the IEEE Conference on Computer Vision and
  Pattern Recognition}, pages 9308--9316, 2019.

\bibitem[Zuo et~al.(2016)Zuo, Ren, Zhang, Gu, and Zhang]{zuo2016learning}
Wangmeng Zuo, Dongwei Ren, David Zhang, Shuhang Gu, and Lei Zhang.
\newblock Learning iteration-wise generalized shrinkage--thresholding operators
  for blind deconvolution.
\newblock \emph{IEEE Transactions on Image Processing}, 25\penalty0
  (4):\penalty0 1751--1764, 2016.

\end{thebibliography}
\end{document}